\documentclass{bmvc2k}

\title{Certainty Volume Prediction for Unsupervised Domain Adaptation}

\addauthor{Tobias Ringwald}{tobias.ringwald@kit.edu}{1}
\addauthor{Rainer Stiefelhagen}{rainer.stiefelhagen@kit.edu}{1}

\addinstitution{
 Institute for Anthropomatics and Robotics (CV:HCI Lab)\\
 Karlsruhe Institute of Technology\\
 Karlsruhe, Germany
}

\runninghead{Ringwald \etal}{Certainty Volume Prediction for UDA}
\def\eg{\emph{e.g}\bmvaOneDot}

\def\etal{\emph{et al}\bmvaOneDot}

\usepackage{bbold}

\usepackage{epsfig}
\usepackage{graphicx}
\usepackage{amsmath}
\usepackage{amssymb}

\usepackage{placeins}

\usepackage{bbding}
\usepackage{pifont}
\usepackage{wasysym}
\usepackage{amsmath}
\usepackage{mathtools}
\usepackage{amssymb}
\usepackage{array}
\usepackage{arydshln}
\usepackage{multirow}
\usepackage[ruled,vlined,noend]{algorithm2e}
\usepackage{booktabs}
\usepackage{adjustbox}
\usepackage{color, colortbl}

\newcommand\Tstrut{\rule{0pt}{2.3ex}}
\newcommand{\z}[2]{#1{\small $\pm$#2}}
\def\wrt{w.r.t\bmvaOneDot}

\usepackage{cellspace}
\begin{document}

\maketitle

\begin{abstract}
Unsupervised domain adaptation (UDA) deals with the problem of classifying unlabeled target domain data while labeled data is only available for a different source domain. Unfortunately, commonly used classification methods cannot fulfill this task adequately due to the domain gap between the source and target data. 
In this paper, we propose a novel uncertainty-aware domain adaptation setup that models uncertainty as a multivariate Gaussian distribution in feature space.
We show that our proposed uncertainty measure correlates with other common uncertainty quantifications and relates to smoothing the  classifier's decision boundary, therefore improving the generalization capabilities.
We evaluate our proposed pipeline on challenging UDA datasets and achieve state-of-the-art results. Code for our method is available at \url{https://gitlab.com/tringwald/cvp}.

\end{abstract}

\section{Introduction}\label{sec:intro}

In recent years, deep neural classification networks have been successfully applied to a wide variety of tasks. One key contribution to that success is the utilization of enormous amounts of labeled training data during the optimization process of such networks. While abundant labeled data might be readily available for a given source domain, more specialized data is often hard to obtain or too expensive to manually label for a desired target domain (\eg medical data). Thus, past research \cite{zou2020game,richter2016playing,ringwald2021adaptiope,visda2017} proposed to generate synthetic data for the training of classifiers, as this would render the manual labeling process unnecessary. Unfortunately, current classification architectures cannot handle the domain shift between the source and target domain, leading to a significantly degraded classification accuracy.

Unsupervised domain adaptation (UDA) seeks to address this domain shift problem by adapting a model to the desired target domain while only given labeled data of the source domain. Recent work has tackled this problem from multiple different angles: Holistic image-level and pixel-level methods often try to manipulate the global appearance of a given source or target domain image, for example by applying style transfer or other image-to-image mapping algorithms \cite{hoffman2017cycada,atapour2018real}. At the feature-level, prior methods commonly focus on minimizing domain divergence measures \cite{can,meng2018adversarial} or enforce source and target features to be indistinguishable through the means of adversarial training \cite{ganin2014unsupervised,simnet}. 

However, these methods often neglect the intrinsic uncertainty accompanying the unsupervised domain adaptation process: as no target domain labels are available, pseudo-labels have to be constructed, which is an inherently noisy task involving uncertainty. In this paper, we thus propose a novel feature-level unsupervised domain adaptation algorithm called Certainty Volume Prediction (CVP). CVP integrates uncertainty into the domain adaptation pipeline by modelling a multivariate Gaussian distribution around a given feature. The covariance matrix of said distribution is constructed in a way that reflects the model's current certainty (or uncertainty) for a feature given the current training step. We achieve this by enforcing samples from the distribution to belong to the same class as the original feature. Therefore, features for which the model is more certain can be allotted higher variance values than uncertain features, due to their distance from the decision boundary.

Finally, we show that our proposed feature space method correlates with other common uncertainty measures in classifier space -- such as Monte-Carlo Dropout~\cite{mcdropout} -- and relates to smoothing the decision boundary \cite{Bartlett98generalizationperformance,lee1995lower,rodriguez2020embedding} which has been linked to better generalization and robustness properties~\cite{Hindi2011SmoothingDB}. In summary, our contributions are threefold:
(i) We introduce our novel Certainty Volume Prediction algorithm for measuring uncertainty in feature space and show how it can help to improve a model's domain adaptation capabilities. (ii) We show its correlation to other uncertainty measures such as Monte-Carlo Dropout. (iii) We show CVP's state-of-the-art performance on multiple challenging datasets and link its increased generalization performance to the smoothness of the decision boundary.

\begin{figure}[t!]
\centering
\vspace{0.25cm}
\begin{tabular}{c}
\bmvaHangBox{%
\centering
\includegraphics[width=0.9\textwidth]{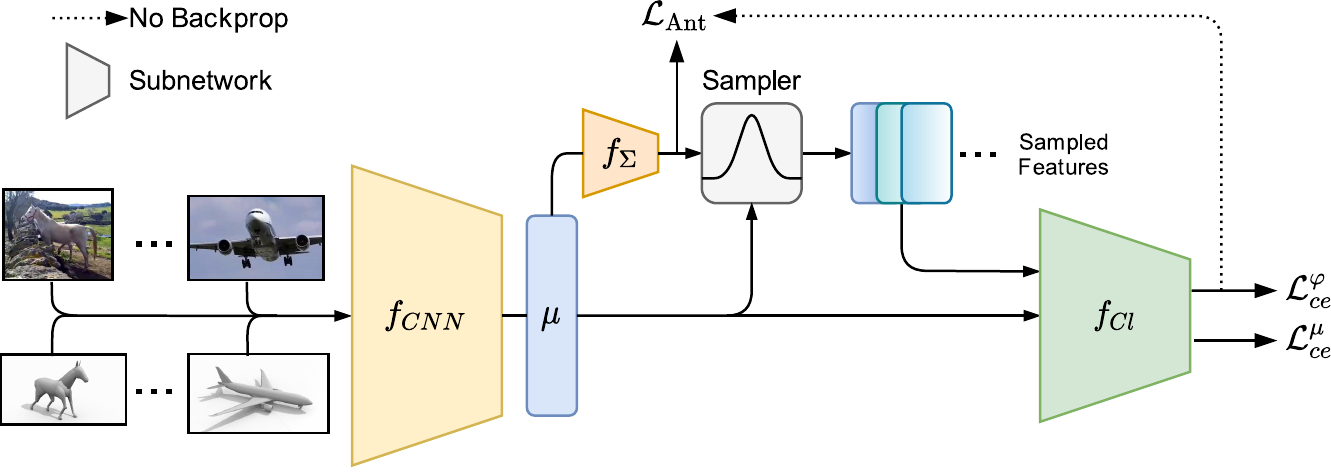}%
}
\end{tabular}
\vspace{0.25cm}
\caption{Overview of our proposed Certainty Volume Prediction (CVP) architecture. The predicted feature mean and covariance matrix predicted by the $f_\Sigma$ subnetwork are forwarded to the feature sampler, that draws additional samples according to the distribution. These samples and the original mean features are classified by the $f_\text{Cl}$ subnetwork and used for our proposed loss functions.}
\label{fig:overview_fig}
\end{figure}

\section{Related Work}\label{sec:related}

Recent works in the area of domain adaptation have viewed the domain shift problem from multiple different angles: At a holistic image-level (often referred to as pixel-level), methods often try to find a common input or latent space in order to reuse the same classifier for both source and target domain. This can be achieved by the means of style transfer \cite{atapour2018real} and related image-to-image translation methods \cite{hoffman2017cycada,murez2018image}. A domain-agnostic space can also be created at the feature-level instead of the image-level. For this, Ganin~\etal propose their gradient reversal layer (RevGrad) in conjunction with a domain classifier, which enforces features from the source and target domain to be indistinguishable. 
Additionally, distribution divergence measures have been applied at the feature-level for the alignment of the source and target domain distributions: Long~\etal~\cite{long2013transfer} employ a maximum mean discrepancy (MMD)~\cite{gretton2006kernel_mmd} based method in order to jointly adapt the marginal and conditional distributions. This idea of MMD-based optimization has recently been picked up again by Kang~\etal~\cite{can} for their proposed CDD loss and is also used for regularization in the PUnDA framework by Gholami~\etal~\cite{gholami2017punda}.
Motivated by the results of Szegedy and Goodfellow~\cite{szegedy2013intriguing,goodfellowgenerative}, adversarial approaches have also been investigated. In this subarea, Park~\etal~\cite{park2018adversarial} have applied adversarial dropout to improve generalization while Pinheiro~\cite{simnet} proposes a categorical prototype-based domain-adversarial learning setup. 
Another orthogonal research directions was pursued by Chang~\etal~\cite{dsbn}, who aim at capturing the different source and target distributions in domain-specific batch-normalization layers.

In this work, we focus on the integration of uncertainty into the domain adaptation pipeline. Related to this topic, Han~\etal~\cite{han2019unsupervised} proposed to calibrate the predictive uncertainty on the target domain data through the source domain uncertainties. They model this uncertainty in the classifier space and quantify it as the predicted Rényi entropy on the target domain data.
Guan~\etal~\cite{guan2021uncertainty} propose the UaDAN framework that employs conditional adversarial learning for alignment of samples. They measure uncertainty as a proxy metric based on the alignment quality of samples and use this in the alignment process as well as their proposed curriculum learning setup.
Furthermore, Ringwald~\etal~\cite{ringwald2020unsupervised} propose the UFAL architecture, which measures uncertainty through Monte-Carlo Dropout (MCD) in classifier space. They then utilize this uncertainty for alignment of features in an Euclidean feature space and also for a curriculum learning-like filtering step.

Unlike the aforementioned approaches, our Certainty Volume Prediction (CVP) framework does not use post hoc measures for uncertainty but instead directly models uncertainty as a multivariate normal distribution in feature space, which is integrated in the backpropagation path of our model and can therefore be optimized during training.
We show that our proposed uncertainty measure in feature space correlates with commonly used uncertainty measures in classifier space (such as Monte-Carlo Dropout) and can also be interpreted with regard to intuitive key assumptions. Finally, we show that CVP's improved generalization performance can be linked to the smoothness of the classifier's decision boundary.

\section{Methodology}\label{sec:methodology}

\subsection{Basic Setup and Naming Conventions} 

In an unsupervised domain adaptation (UDA) setup, labeled instances are only available for the source domain while the target domain purely consists of unlabeled data. Formally, we consider the source domain dataset $\mathcal{D}_S = \{(x_S^{(i)}, y_S^{(i)})\}_{i=1}^{N_S}$ to contain $N_S$ pairs of images $x_S^{(i)}$ and labels $y_S^{(i)} \in \mathcal{C}$ where $\mathcal{C}=\{1, 2,\ldots, N_C\}$ is the set of classes with size $\lvert \mathcal{C} \rvert = N_C$, while the target domain dataset $\mathcal{D}_T = \{(x_T^{(i)})\}_{i=1}^{N_T}$ only consists of $N_T$ unlabeled instances. The objective of an UDA algorithm is then to produce accurate predictions $y_T^{(i)}$ for every target domain instance $x_T^{(i)}$.

The UDA approach presented in this paper is based on a common convolutional neural network (CNN) feature extractor, which we denote as $f_\text{CNN}$. The extracted features are subsequently processed by our proposed feature sampling stage and a classifier $f_\text{Cl}$, which maps its input into a probability distribution over the class set $\mathcal{C}$. A basic overview of the full architecture is provided in Figure~\ref{fig:overview_fig}. Given these prerequisites, we will introduce our full CVP architecture in the following sections.

\subsection{Advantages of Feature Space Uncertainty Estimation}

The goal of our proposed Certainty Volume Prediction method is to model uncertainty in feature space in a way that improves the generalization process necessary for domain adaptation. This is in contrast to prior approaches \cite{ringwald2020unsupervised,han2019unsupervised,mcdropout} that model and measure uncertainty in the classifier-space. For example, Monte Carlo dropout \cite{mcdropout} proposes to conduct $T$ forward passes at testing time using different dropout~\cite{dropout} masks. This results in $T$ different probability distributions over the class set $\mathcal{C}$ which can then be used to calculate a model's predictive mean and variance. Both of these values can be interpreted as a form of uncertainty measure. However, we argue that uncertainty should be estimated earlier in a model's pipeline, as the classifier output has already been reduced to a much smaller, less informative space through mapping into class-space over $\mathcal{C}$ and squashing by softmax normalization. Thus, we propose to use the feature space for the uncertainty estimation.

\subsection{Modeling Uncertainty in Feature Space with CVP}

CNN feature extractors provide a means to map any input image to a compact feature representation. One key property of this mapping process is that similar inputs should be mapped to similar outputs, i.e.\ images of the same class are mapped to features that have a small distance in feature space. This property is commonly exploited by unsupervised algorithms such as clustering and can also be observed in high-dimensional spaces through dimension reduction techniques~(cf. \cite{maaten2008visualizing}). 
Hence, it is often referred to as cluster assumption~\cite{chapelle2005semi} and states that decision boundaries should not cross high density regions.

\begin{figure}[t!]
\centering
\vspace{0.25cm}
\begin{tabular}{ccc}
\bmvaHangBox{\centering\includegraphics[width=0.4\textwidth]{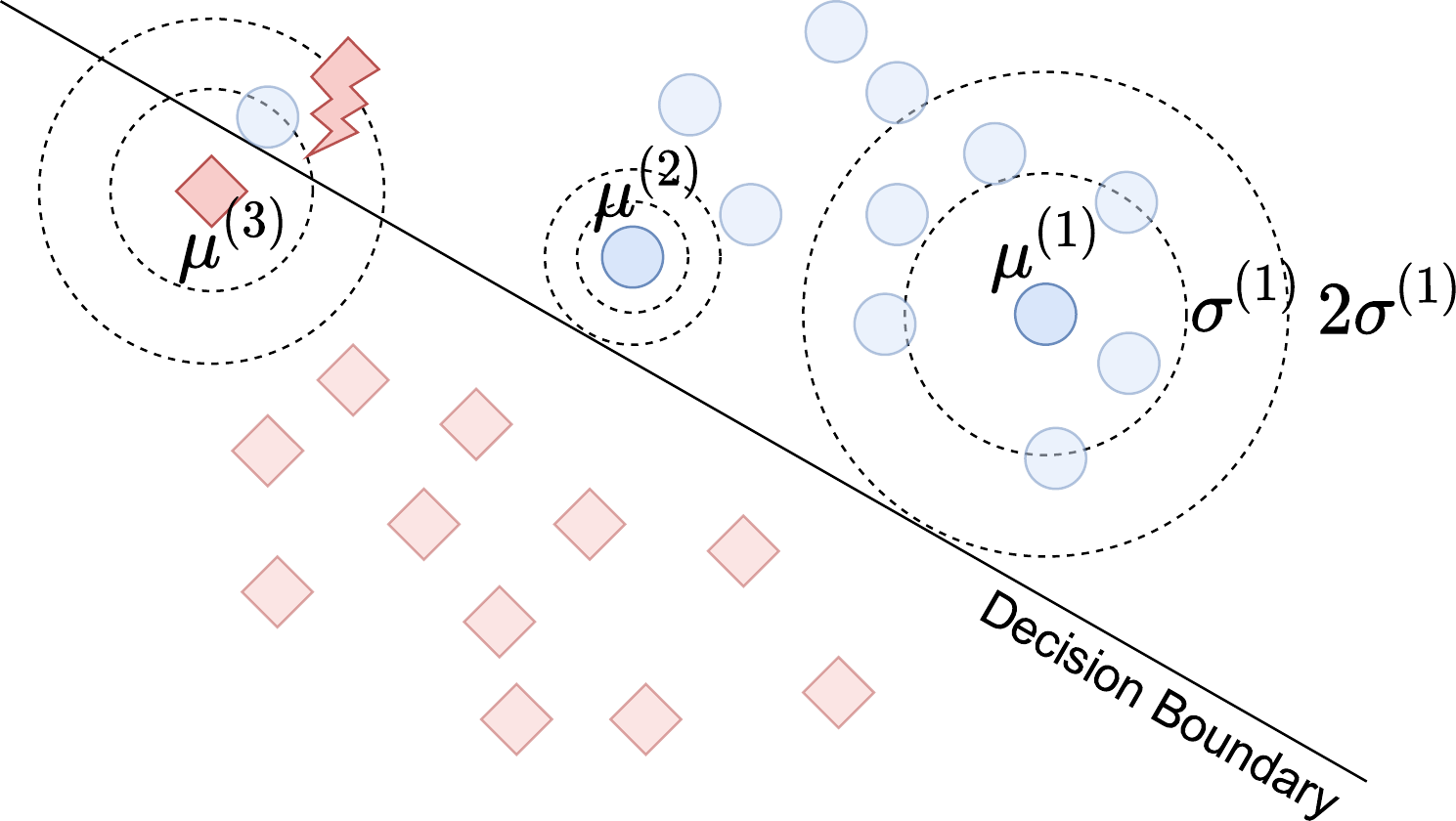}} & \bmvaHangBox{\centering\includegraphics[width=0.25\textwidth]{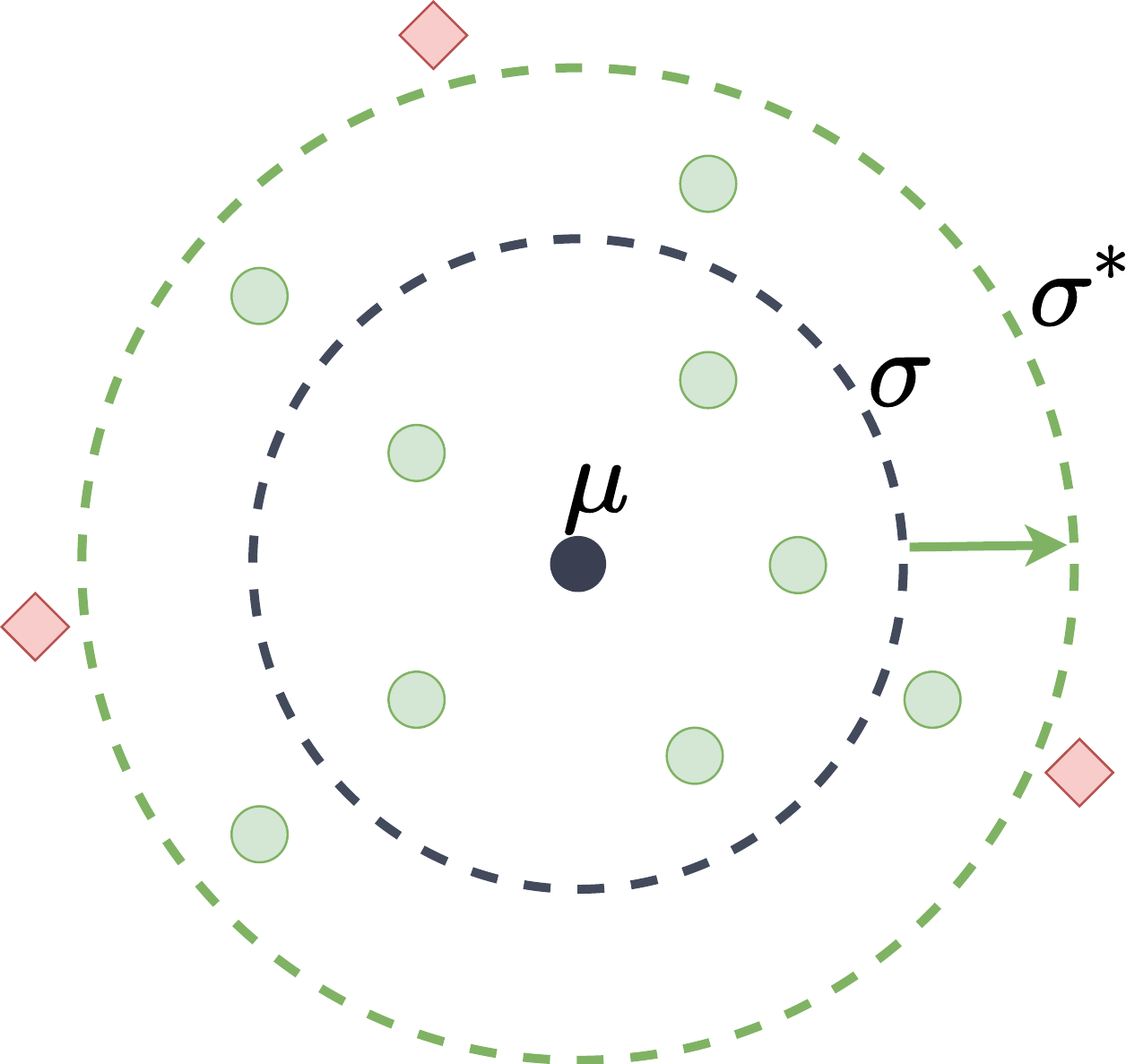}} & \bmvaHangBox{\centering\includegraphics[width=0.25\textwidth]{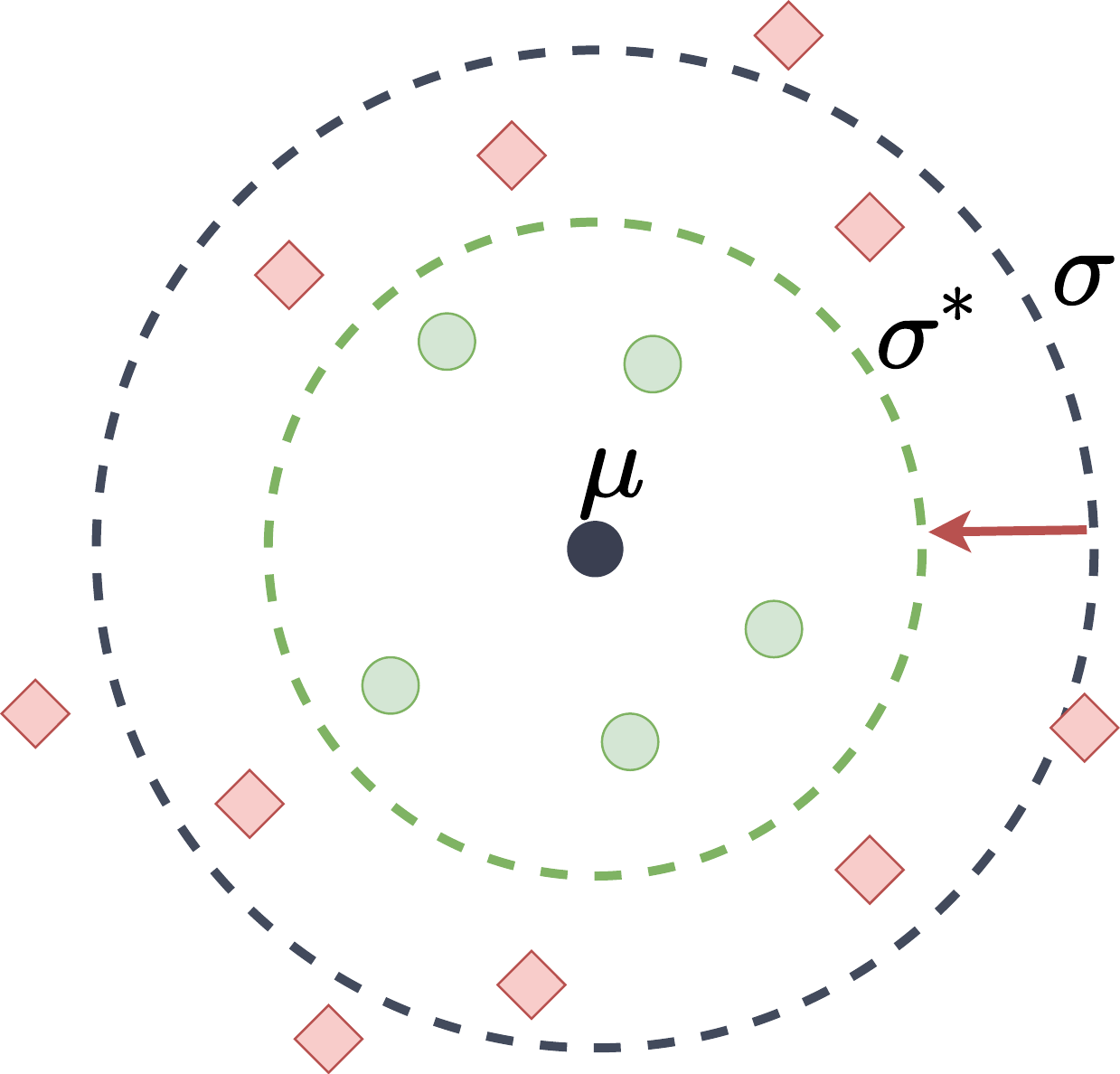}}\\
(a) & (b) & (c)\\
\end{tabular}
\vspace{0.25cm}
\caption{From left to right: (a) Visualization of CVP's interaction with the decision boundary. (b) A large amount of correctly classified instances will result in a small $\mathcal{L}_{ce}^\varphi$ value and therefore an increase of $\sigma$. (c) Likewise, a large amount of incorrect classifications will increase $\mathcal{L}_{ce}^\varphi$ and lower the training target $\psi$ and hence also $\sigma$.}
\label{fig:trainofthought}
\end{figure}

Our CVP algorithm also utilizes this property: We first apply a feature extractor $f_\text{CNN}$ to a given input image $x^{(i)}$ in order to compute a compact feature embedding $\mu^{(i)}$. 
Given above cluster assumption, features near $\mu^{(i)}$ should belong to the same class as $\mu^{(i)}$ itself. 
However, this leads to the question how to define what ``near'' means and how to obtain these features. In this paper, we propose to phrase proximity as an uncertainty estimation in feature space based on the decision boundary and visualize this in Figure~\ref{fig:trainofthought}. To achieve this, we model a multivariate Gaussian distribution around the predicted mean $\mu^{(i)}$. We employ a subnetwork $f_\Sigma$ (see Figure~\ref{fig:overview_fig}) to predict a suitable covariance matrix $\Sigma^{(i)}$, which enables us to sample related features from the resulting distribution $\mathcal{N}(\mu^{(i)}, \Sigma^{(i)})$.
Given a $D$-dimensional $\mu$ feature, the matching covariance matrix would be of shape $D\times D$. However, for common CNN feature extractors, $D$ is often very large (\eg $D=2048$ for ResNet-50~\cite{he2016deep}) which would result in a covariance matrix with millions of values and an even larger amount of weights in $f_\Sigma$ to predict those values. We thus propose to use a computationally feasible subnetwork $f_\Sigma: \mu^{(i)} \in \mathbb{R}^D \rightarrow \sigma^{(i)} \in \mathbb{R}$ and then sample from the distribution $\mathcal{N}(\mu^{(i)}, (\sigma^{(i)}\mathbb{1})^2)$ where $\mathbb{1}$ denotes the identity matrix.

Intuitively, features far from a decision boundary are located in an area of high certainty, i.e.\ many of the surrounding feature belong to the same class. We visualize this in Figure~\ref{fig:trainofthought}a for a 2-dimensional case: the distribution centered on $\mu^{(1)}$ can be assigned a large $\sigma$ without crossing a decision boundary, as it is located in such an area of high certainty. 
Features very close to the decision boundary must not have a large $\sigma$ (see $\mu^{(2)}$) as this might cross the decision boundary and violates above assumption that nearby features should belong to the same classes (see $\mu^{(3)}$). Intuitively, it follows that $\sigma$ can be interpreted as an uncertainty estimate for a given input, where small $\sigma$ values signal uncertainty while large $\sigma$ values indicate certainty. The objective of our Certainty Volume Prediction setup is then to assess what area -- or volume in higher dimensions -- can be safely spanned by the distribution so that the majority of samples will not cross a decision boundary.

\subsection{Optimization of CVP}\label{sec:optim_of_cvp}

We meet CVP's objective defined above with the help of two loss functions. After obtaining $\mu^{(i)}$ and $\sigma^{(i)}$ given the above description, we draw $M$ feature samples from the resulting distribution $\varphi_{1..M}^{(i)}\sim\mathcal{N}(\mu^{(i)}, (\sigma^{(i)}\mathbb{1})^2)$. In order to satisfy the cluster assumption, we enforce all of these new features to belong to the same class $y^{(i)}$ as $\mu^{(i)}$. For this, we use the cross-entropy-based samples loss $\mathcal{L}_{ce}^\varphi$ (see Equation~\ref{eq:samples_xent}) where $\hat{y}^{(i)}$ is the one-hot encoded label corresponding to $y^{(i)}$. Applying this loss will contract $\sigma^{(i)}$ if the certainty estimation is too high, i.e.\ samples are drawn too far from the mean and thus misclassified which increases the cross-entropy term. However, using the $\mathcal{L}_{ce}^\varphi$ loss alone would collapse $\sigma^{(i)}$ towards 0 as this would minimize the objective function. Therefore, we also propose an additional antagonistic loss $\mathcal{L}_{Ant}$ (see Equation~\ref{eq:ant_loss}). $\mathcal{L}_{Ant}$ uses the $\mathcal{L}_{ce}^\varphi$-based variable $\psi^{(i)}$ (see Equation~\ref{eq:psi}) in order to enforce an inversely proportional relationship between the samples loss $\mathcal{L}_{ce}^\varphi$ and the (un-)certainty estimation $\sigma^{(i)}$: The higher $\mathcal{L}_{ce}^\varphi$ becomes, the lower $\sigma^{(i)}$ should be; at the maximum expected loss value for $\mathcal{L}_{ce}^\varphi$ (denoted as $\kappa$), the target value $\psi^{(i)}$ for $\sigma^{(i)}$ should be 0. We finally model this as a regression task utilizing a smooth $\ell_1$ loss between $\sigma^{(i)}$ and $\psi^{(i)}$ (see Equation~\ref{eq:ant_loss}). We provide a theorem on how to derive the constant $\kappa$ in Appendix~\ref{sec:kappa}.

\begin{align}
    \mathcal{L}_{ce}^{\varphi}(\hat{y}^{(i)}, \varphi^{(i)}) = -\frac{1}{M}\sum_{m=1}^{M} \sum_{j=1}^{\lvert \mathcal{C} \rvert} \hat{y}^{(i)}_j\; \mathrm{log}(f_\text{Cl}(\varphi^{(i)}_m))_j \label{eq:samples_xent}
\end{align}

\begin{align}
    \psi^{(i)}(\hat{y}^{(i)}, \varphi^{(i)}) = \mathrm{max}(\{0,\; \kappa - \mathcal{L}_{ce}^{\varphi}(\hat{y}^{(i)}, \varphi^{(i)})\}) \label{eq:psi}
\end{align}

\begin{align}
    \mathcal{L}_{Ant}(\sigma^{(i)}, \hat{y}^{(i)}, \varphi^{(i)}) = \begin{cases}
      \frac{1}{2} (\sigma^{(i)}-\psi^{(i)})^2  & \text{if}\; \lvert \sigma^{(i)}-\psi^{(i)} \rvert < 1\\
       \lvert \sigma^{(i)}-\psi^{(i)} \rvert & \text{otherwise}\\
    \end{cases} \label{eq:ant_loss}
\end{align}

Finally, we add the last loss term $\mathcal{L}_{ce}^{\mu} = -\sum_{j=1}^{\lvert \mathcal{C} \rvert} \hat{y}^{(i)}_j\; \mathrm{log}(f_\text{Cl}(\mu^{(i)}))_j$, which is equivalent to the normal feature classification in a standard CNN training setup. We thus arrive at the total loss $\mathcal{L}_{total} = \mathcal{L}_{ce}^{\mu} + \alpha \mathcal{L}_{ce}^{\varphi} + \mathcal{L}_{Ant}$ where $\alpha$ is a tuneable loss weight.

\subsection{Training Process}

Given above formal introduction of our proposed CVP algorithm, we can now describe our full domain adaptation pipeline. Overall, our setup is two-stage with an initial pretraining phase and an iterative domain adaptation refinement step.
\newline
\textbf{Pretraining Phase}. Our training setup starts with a pretraining phase. During this phase, only labeled source domain data is used. This phase mainly serves the purpose of initializing the weights of the $f_\Sigma$ and $f_\text{Cl}$, which can be reliably done with the ground truth source domain annotations. At the end of the pretraining phase, the combination of $f_\Sigma(f_\text{CNN}(\cdot))$ can already be used to estimate uncertainty while $f_\text{Cl}(f_\text{CNN}(\cdot))$ can be employed as a pseudo-labeler.
\newline
\textbf{Adaptation Phase}. The adaptation phase is an iterative cycle-based refinement step, during which the model is adapted to the target domain. At the start of every adaptation cycle, we utilize the current training progress of our model in order to create pseudo-labels for target domain instances, i.e.\ we use $\mathit{argmax} \circ f_\text{Cl} \circ f_\text{CNN}: x_T^{(i)} \rightarrow y_T^{(i)}$ to generate a label $y_T^{(i)} \in \mathcal{C}$ for every target domain instance $x_T^{(i)}$. We then follow the procedure of~\cite{ringwald2020unsupervised} and sample mixed domain batches consisting of 50\% source domain and 50\% target domain images for training. Given such a batch, the model is then optimized \wrt the loss function $\mathcal{L}_\text{total}$ introduced in section~\ref{sec:optim_of_cvp}. This is repeated for $T_\text{cycle}$ steps, until the next adaptation cycle begins.
The pseudo-label generation always uses the current model weights, therefore only one set of model weights needs to be kept, i.e.\ no additional memory or storage space is necessary.

\section{Experiments}\label{sec:experiments}

\subsection{Evaluation Setup} \label{sec:setup}

\begin{figure}[t!]
\renewcommand{\arraystretch}{2.0}
\begin{tabular}{ccc}
\bmvaHangBox{
	\begin{minipage}[c][1\width]{0.25\textwidth}
	   \centering
        \frame{\includegraphics[width=0.3\textwidth]{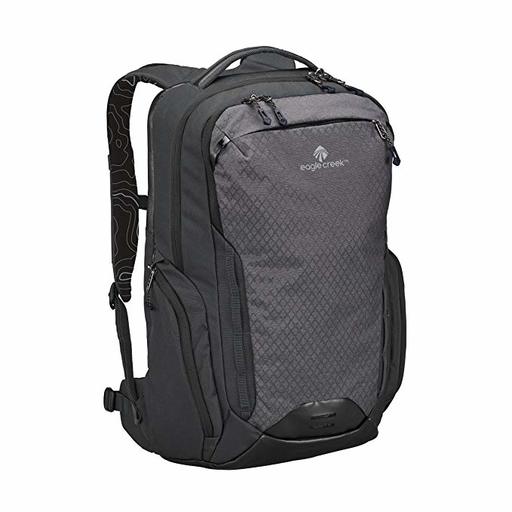}}\hspace{1mm}%
        \frame{\includegraphics[width=0.3\textwidth]{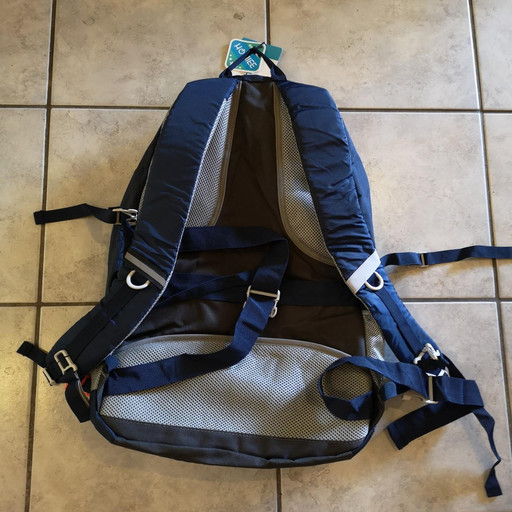}}\hspace{1mm}%
        \frame{\includegraphics[width=0.3\textwidth]{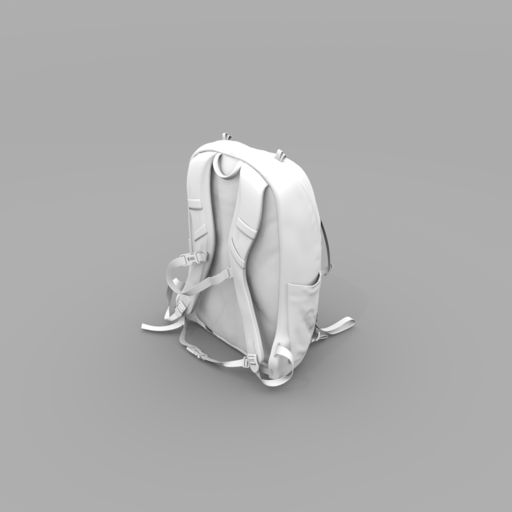}}\hspace{1mm}\\\vspace{0.5mm}
        \frame{\includegraphics[width=0.3\textwidth]{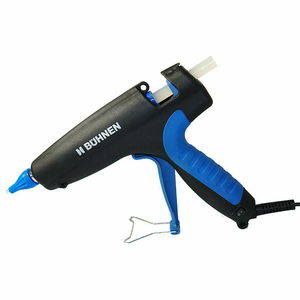}}\hspace{1mm}%
        \frame{\includegraphics[width=0.3\textwidth]{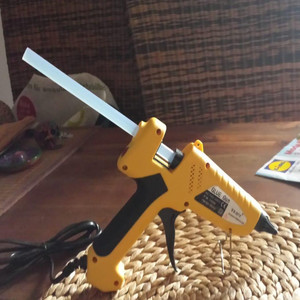}}\hspace{1mm}%
        \frame{\includegraphics[width=0.3\textwidth]{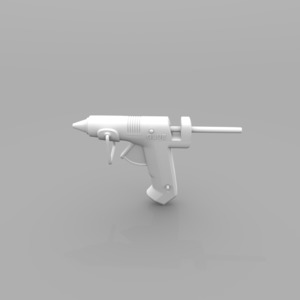}}\\\vspace{0.5mm}
        \frame{\includegraphics[width=0.3\textwidth]{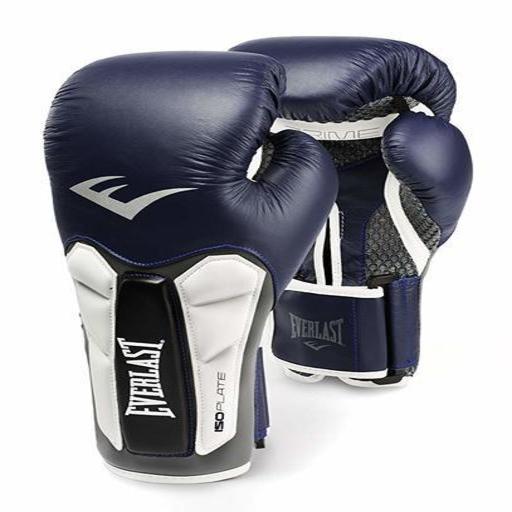}}\hspace{1mm}%
        \frame{\includegraphics[width=0.3\textwidth]{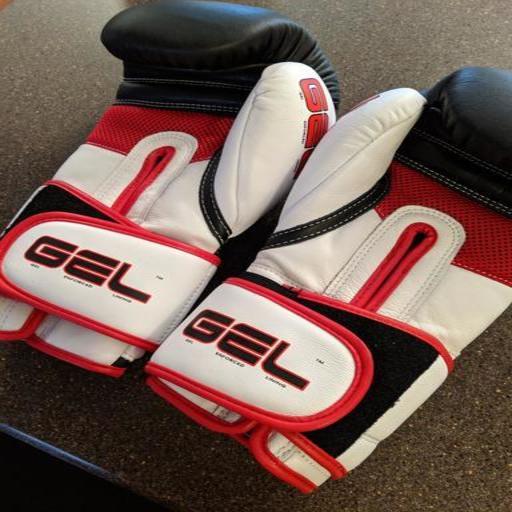}}\hspace{1mm}%
        \frame{\includegraphics[width=0.3\textwidth]{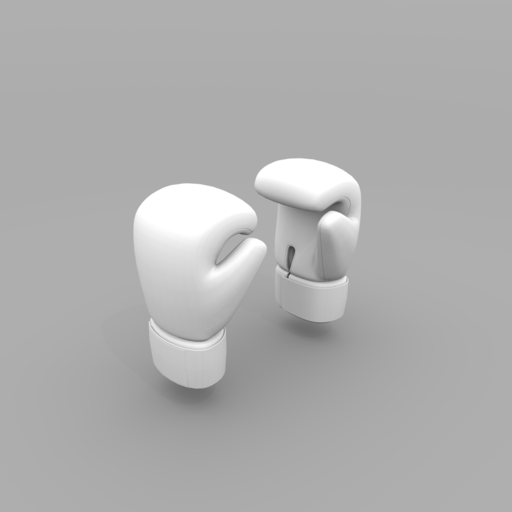}}%
	\end{minipage}} & \bmvaHangBox{
	\begin{minipage}[c][1\width]{0.25\textwidth}
	   \centering
        \frame{\includegraphics[width=0.3\textwidth]{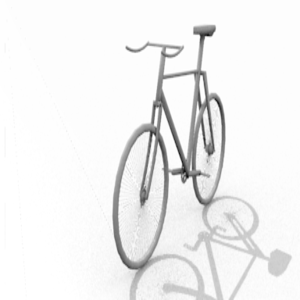}}\hspace{1mm}%
        \frame{\includegraphics[width=0.3\textwidth]{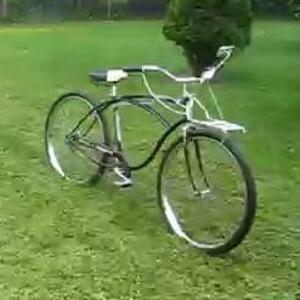}}\hspace{1mm}\\\vspace{0.5mm}
        \frame{\includegraphics[width=0.3\textwidth]{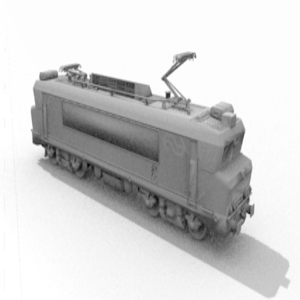}}\hspace{1mm}%
        \frame{\includegraphics[width=0.3\textwidth]{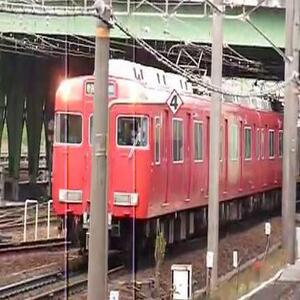}}\hspace{1mm}\\\vspace{0.5mm}
        \frame{\includegraphics[width=0.3\textwidth]{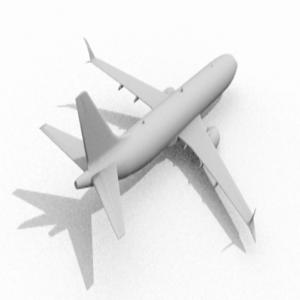}}\hspace{1mm}%
        \frame{\includegraphics[width=0.3\textwidth]{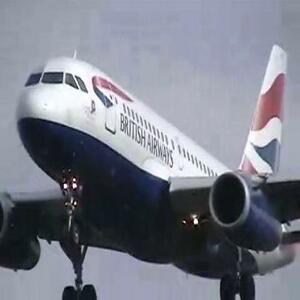}}\hspace{1mm}\\\vspace{0.5mm}
	\end{minipage}} &  \bmvaHangBox{\frame{\includegraphics[width=0.36\textwidth]{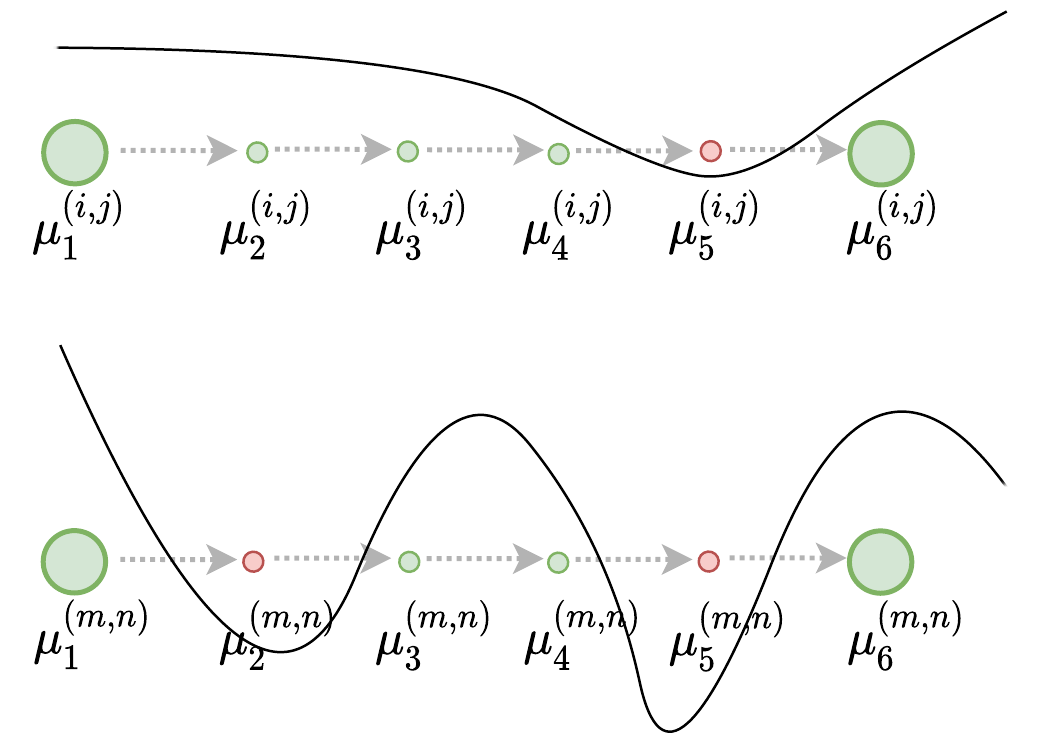}}}\\
	(a) Adaptiope & (b) VisDA-2017 & (c) Boundary smoothness evaluation\\
\end{tabular}
\caption{(a) Example images from Adaptiope~\cite{ringwald2021adaptiope} with 123 classes in the domains product, real and synthetic. (b) Example images from VisDA-2017~\cite{visda2017} with 12 classes in the domains synthetic and real. (c) Visualization of the boundary smoothness evaluation. Top: Interpolating between $\mu^{(i)}$ and $\mu^{(j)}$ results in few class changes due to the smooth decision boundary. Bottom: Interpolating between $\mu^{(m)}$ and $\mu^{(n)}$ results in many changes due to the curvy shape of the boundary.} 
\label{fig:datasets}
\end{figure}

\textbf{Datasets}. We evaluate our proposed CVP method on four publicly available UDA benchmark datasets: 
\textit{VisDA-2017} (also known as Syn2Real-C)~\cite{visda2017} is a staple dataset for unsupervised domain adaptation that offers over 200,000 images from 12 object classes in the domains synthetic and real. The synthetic domain (train set) contains 152,397 images of 3D renderings under different lighting conditions and viewing angles. The real domain (test set) contains 72,372 real life images extracted from the YouTube Bounding-Boxes dataset~\cite{real2017youtube}. 
\textit{Adaptiope}~\cite{ringwald2021adaptiope} is a very challenging UDA dataset with 123 classes in the three domains synthetic, product and real life. Adaptiope was constructed as a balanced dataset. Therefore, every class and every domain contains the same amount of images, resulting in 36,900 images overall with 100 images per class and 12,300 images per domain.
\textit{Office-Caltech}~\cite{officecaltech} is a common UDA dataset containing 2,533 images of 10 classes in the 4 domains Amazon, Caltech, DSLR and Webcam. \textit{Modern Office-31}~\cite{ringwald2021adaptiope} was recently proposed as a replacement for the Office-31~\cite{office31} dataset. It rectifies the annotation errors in Office-31 and also adds a new synthetic domain for a total of 6,712 images of 31 classes in the 3 domains Amazon, Webcam and Synthetic.
Example images from the datasets are shown in Figure~\ref{fig:datasets}.
\newline
\textbf{Evaluation Metrics}. For the evaluation of our method, we employ the standard metrics of each dataset: For Adaptiope, we construct six transfer tasks based on the three domains and calculate a mean and standard deviation over multiple runs as proposed in~\cite{ringwald2021adaptiope}. For VisDA-2017, we calculate the mean class accuracy as per challenge evaluation protocol. Here, the standard accuracy is computed per class and then averaged over all 12 classes for a final score. For Modern Office-31 and Office-Caltech, we construct 6 and 12 transfer tasks respectively and report the accuracy per task as well as an overall average.
For the evaluation of the decision boundary smoothness, we employ the \textit{oscillation of classification} metric from~\cite{karimi2019characterizing}. Here, we generate subfeatures by linearly interpolating between two features $\mu^{(i)}$ and $\mu^{(j)}$ (denoted as $\mu^{(i,j)}_k$ with $k$ being the interpolation step) and measure the number of changes in classification. Formally, we measure 
$\mathcal{O}(\mu^{(i)}, \mu^{(j)})= \frac{1}{K}\sum_{k=1}^{K-1} \mathbf{1}\left(\mathrm{argmax}\; f_\text{Cl}(\mu^{(i,j)}_k) \neq \mathrm{argmax}\; f_\text{Cl}(\mu^{(i,j)}_{k+1})\right)$ where $\mathbf{1}$ denotes the indicator function and $K$ is the number of interpolation steps. 
This metric is visualized in Figure~\ref{fig:datasets}c for $K=6$.
We provide further information about the implementation details and the evaluation setup in Appendix~\ref{sec:appendix_implementation}.

\subsection{Results}

\begin{figure}[!t]
\renewcommand{\arraystretch}{2.0}
\begin{tabular}{ccc}
\bmvaHangBox{\includegraphics[width=0.3\textwidth]{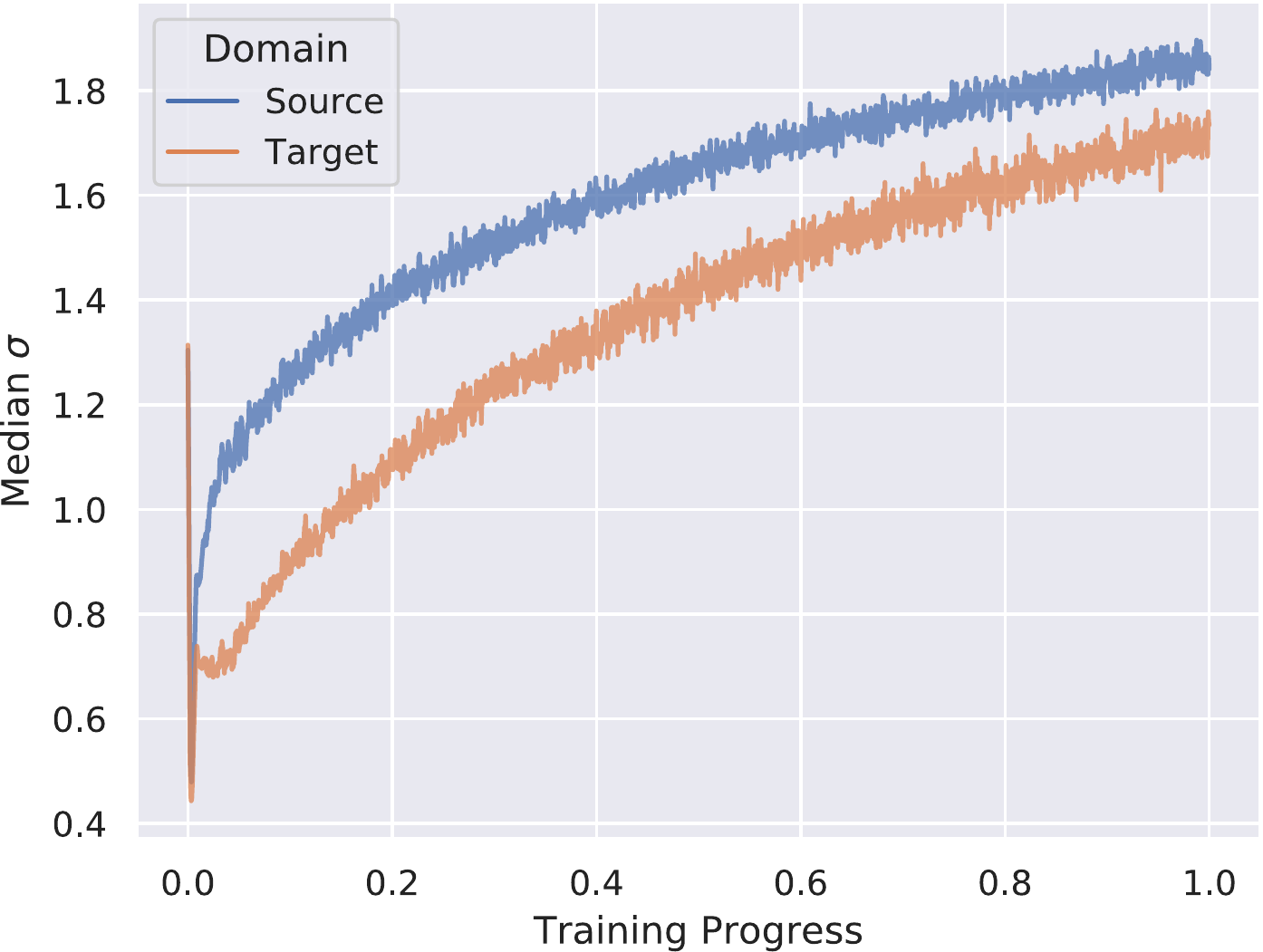}} & 
\bmvaHangBox{\includegraphics[width=0.3\textwidth]{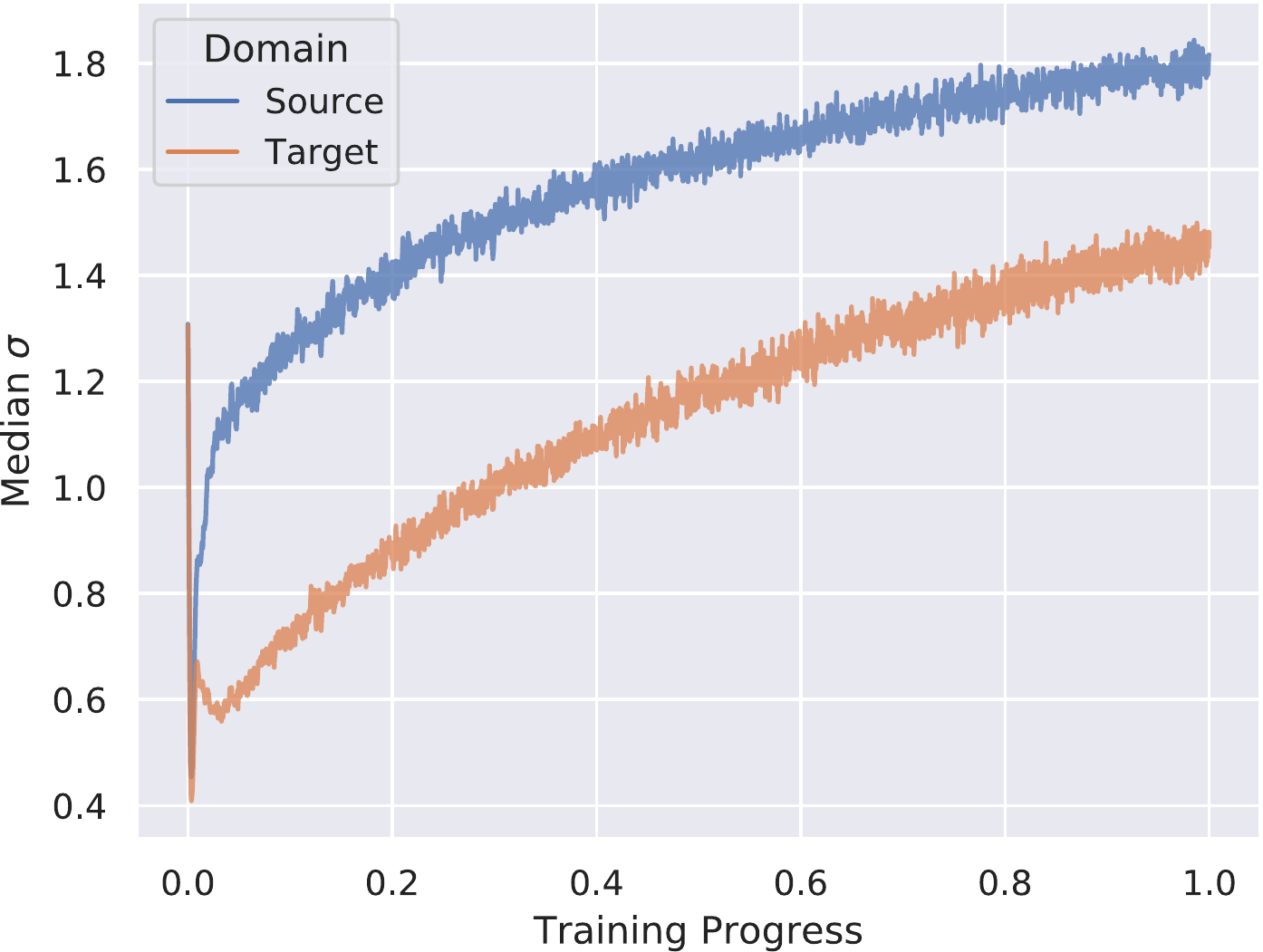}} &
\bmvaHangBox{\includegraphics[width=0.3\textwidth]{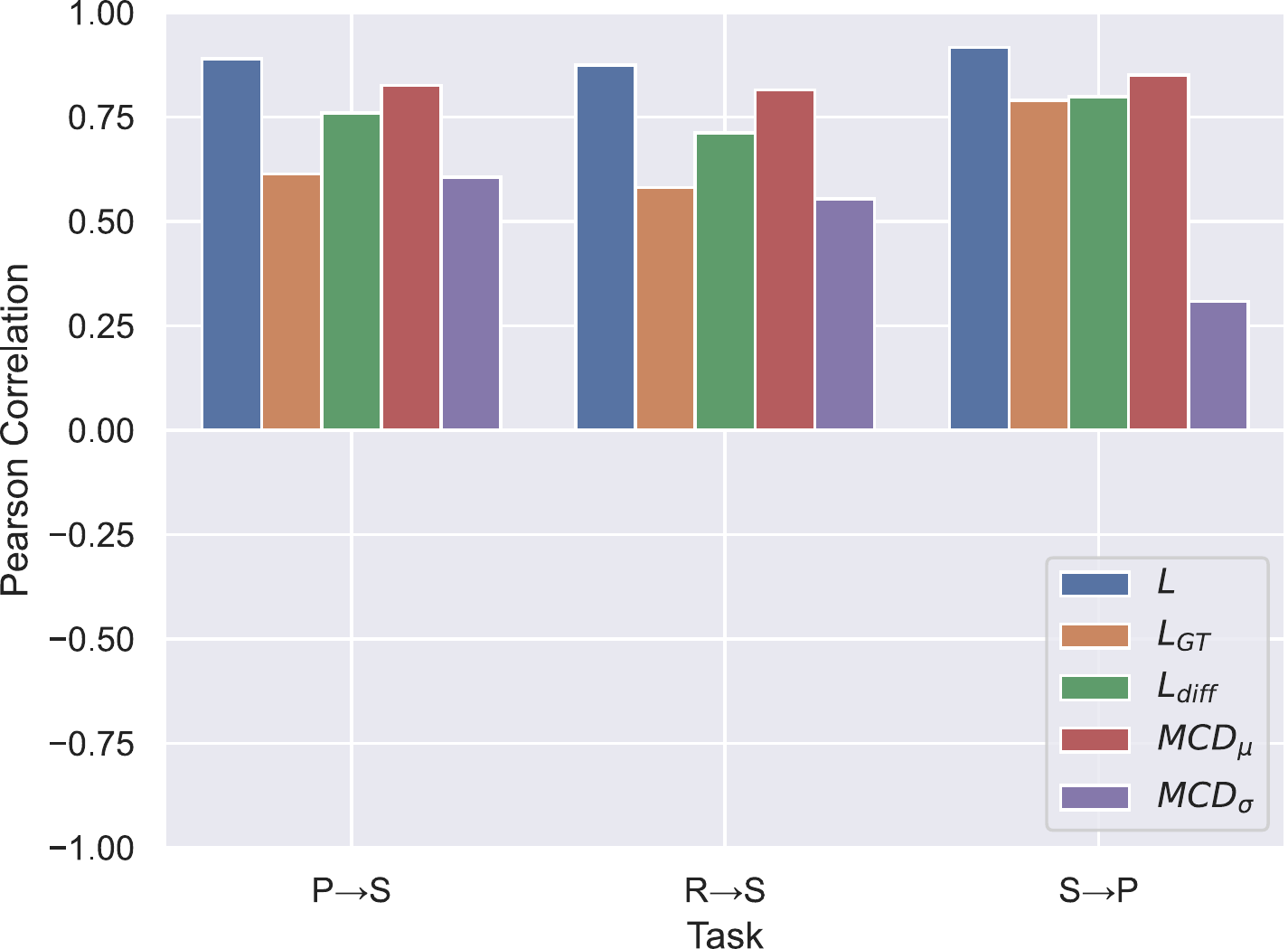}} \\
(a) S$\rightarrow$P & (b) S$\rightarrow$R & (c) Correlation\\
\end{tabular}
\caption{(a, b) Median $\sigma$ values for the source and target domain as the training progresses. (c) Correlation of $\sigma$ with common uncertainty measures. Best viewed in the digital version.} 
\label{fig:analysis}
\end{figure}

\textbf{Analysis}. We start evaluating our CVP method by providing an analysis \wrt CVP's $\sigma$ prediction. In Figure~\ref{fig:analysis}a and b, we show the batch median $\sigma$ for both source and target domain samples during the adaptation phase as training progresses. During the pretraining phase, the model has only been trained on source domain data. For this reason, a large drop of $\sigma$ can be observed at the beginning of the adaptation phase, as the model is now confronted with target domain data for the first time. The resulting decrease in $\sigma$ can be interpreted as a lower certainty in the model's predictions due to the involvement of new and unseen data. While the certainty on the source domain quickly recovers, the target domain certainty increase is delayed due to the domain gap and the ongoing domain adaptation during training. For the S$\rightarrow$P task, our CVP model converges to a similar certainty (i.e.\ magnitude of $\sigma$) for both domains. However, for the hardest transfer task of Adaptiope -- S$\rightarrow$R -- the mean target domain certainty at the end of training is still lower than the source domain certainty due to the enormous domain gap of synthetic to real transfer tasks.

Furthermore, we analyze $\sigma$ \wrt common uncertainty measures. In total, we compare with five common uncertainty measures: (1) magnitude of the maximum logit ($L$), (2) magnitude of the logit for the ground truth class ($L_\text{GT}$), (3) the difference of the top two logits ($L_\text{diff}$), (4) the probability at the maximum logit prediction after averaging over multiple Monte Carlo Dropout (MCD) iterations ($MCD_{\mu}$) and (5) the standard deviation at the maximum logit prediction after multiple MCD iterations ($MCD_{\sigma}$, inverted for visualization purposes). We then calculate the Pearson correlation coefficient between $\sigma$ and measure 1-5 and plot the results in Figure~\ref{fig:analysis}c for multiple transfer tasks. For all measures we compare with, we observe a strong linear correlation indicating that $\sigma$ can indeed be interpreted as an (un-)certainty measure. This is also consistent for all evaluated transfer tasks.

\definecolor{Gray}{gray}{0.95}
\begin{table*}[!t]
\centering
\resizebox{0.75\textwidth}{!}{
\begin{tabular}{llcrrrrr}
\toprule
Metric & Setup & Losses & P$\rightarrow$S & R$\rightarrow$S & S$\rightarrow$P & S$\rightarrow$R & Avg. \\\toprule
\multirow{3}{*}{Accuracy} & Basic & $\mathcal{L}^\mu_\text{ce}$ & 69.4 & 70.1 & 74.8 & 60.0 & 68.6 \\
                          & No Samples CE & $\mathcal{L}^\mu_\text{ce}, \mathcal{L}_\text{ant}$ & 70.7 & 70.8 & 77.1 & 61.9 & 70.1 \\
                          & Full \textbf{CVP} Setup & $\mathcal{L}^\mu_\text{ce}, \mathcal{L}_\text{ant}, \mathcal{L}^\varphi_\text{ce}$ & \textbf{71.9} & \textbf{72.4} & \textbf{78.4} & \textbf{63.2} & \textbf{71.5}\\\hline\Tstrut
\multirow{2}{*}{Oscillation} & Basic & $\mathcal{L}^\mu_\text{ce}$ & \z{205.9}{1.3} & \z{207.2}{0.2} & \z{221.2}{3.8} & \z{259.6}{2.7} & 223.5 \\
                             & Full \textbf{CVP} Setup & $\mathcal{L}^\mu_\text{ce}, \mathcal{L}_\text{ant}, \mathcal{L}^\varphi_\text{ce}$ & \textbf{\z{197.1}{0.6}} & \textbf{\z{199.3}{1.2}} & \textbf{\z{208.9}{1.6}} & \textbf{\z{233.0}{4.8}} & \textbf{209.6} \\
\toprule
\end{tabular}
}
\caption{Effect of CVP on the classification results and the smoothness of the decision boundary quantified by the oscillation of classification metric. Results are given as accuracy (in \%, $\uparrow$ higher is better) and osciallation of classification metric $\mathcal{O}(\cdot, \cdot)$ summed over all sampled feature pairs ($\downarrow$ lower is better).}\label{table:smoothness}
\end{table*}

We also conduct an ablation study \wrt our model's three loss functions and show results in Table~\ref{table:smoothness} for the four difficult transfer tasks involving Adaptiope's synthetic domain. In the basic setup, only $\mathcal{L}^\mu_\text{ce}$ is used. This is the minimal setup in order to train the classifier and obtain pseudo-labels. As expected, this performs the worst of all evaluated setups, as CVP's additional loss functions are not active. We continue by adding $\mathcal{L}_\text{ant}$. This already improves the classification accuracy for all four transfer tasks. However, without the $\mathcal{L}^\varphi_\text{ce}$ loss, the cluster assumption is not enforced, which yields excessive $\sigma$ values and results in lower accuracies than our full CVP method. Only the usage of all three losses -- $\mathcal{L}^\mu_\text{ce}$, $\mathcal{L}^\varphi_\text{ce}$ and $\mathcal{L}_\text{ant}$ -- results in the highest accuracy for all four transfer tasks with an average improvement of almost 3\% over the basic setup. Note that training with $\mathcal{L}^\varphi_\text{ce}$ but without $\mathcal{L}_\text{ant}$ would instantly collapse $\sigma$ towards 0 and break the feature sampling due to the zero standard deviation. We thus do not show results for this setup. This is further discussed in Section~\ref{sec:optim_of_cvp}.

Finally, we analyze the smoothness of the decision boundary. For this analysis, we employ the osciallation of classification metric described in Section~\ref{sec:setup}. After the conclusion of CVP's adaptation phase, we extract features $\mu$ for five instances per class and then calculate the sum of this metric for all distinct pairs $(i, j)$, i.e.\ $\sum_i \sum_j \mathcal{O}(\mu^{(i)}, \mu^{(j)})$. The results are shown in Table~\ref{table:smoothness} as mean and standard deviation over multiple runs. Applying CVP clearly leads to a much smoother decision boundary as quantified by the oscillation metric. For every an\-a\-lysed transfer task, the setup using CVP has a multiple standard deviations lower oscillation value, which is a direct indication of a smoother decision boundary. Overall, applying CVP results in an approximately 14 points lower oscillation metric, which translates to 14,000 fewer changes in classification. 
Smoothness of the decision boundary is also strong indicator for the generalization capabilities of a model. This is especially important in UDA, as the trained model is employed as pseudo-labeler for subsequent training steps -- a better generalization would necessarily lead to better pseudo-labels $y_T^{(i)}$. We therefore also provide domain adaptation results with and without CVP in Table~\ref{table:smoothness}. We observe that every model employing CVP improves in accuracy on the target domain when compared to a model without CVP. On average, CVP improves the domain adaptation results by 2.9\%, with a 3.2\% increase on the hardest transfer task S$\rightarrow$R and up to 3.6\% on S$\rightarrow$P. 
We thus conclude that CVP can be seen as an intuitive uncertainty measure that improves domain adaptation capabilities by smoothing the decision boundary of a classifier. For further ablation studies, please refer to Appendix~\ref{sec:appendix_analysis}.

\begin{table*}[t]
\centering
\begin{adjustbox}{width=0.99\textwidth}
\begin{tabular}{lrrrrrrrrrrrrr}
\toprule
    Method & {aero} & {bicyc} & {bus} & {car} & {horse} & {knife} & {motor} & {person} & {plant} & {skate} & {train} & {truck} & {Avg.}\\\toprule
    Source only & 56.0 & 15.3 & 63.4 & 95.3 & 62.0 & 18.4 & 70.2 & 26.2 & 89.1 & 26.0 & 88.6 & 13.2 & 52.0\\
    SE-152~\cite{french2017self} & 88.4 & 84.8 & 75.1 & 84.1 & 80.0 & 72.6 & 63.6 & 56.6 & 95.4 & 73.8 & 77.6 & 78.3 & 77.5 \\
    BUPT \cite{visda2017} & \textbf{95.7} & 67.0 & \textbf{93.4} & \textbf{97.2} & {90.6} & 86.9 & \textbf{92.0} & 74.2 & {96.3} & 66.9 & \textbf{95.2} & 69.2 & 85.4\\
    CAN \cite{can} & --- & --- & --- & --- & --- & --- & --- & --- & --- & --- & --- & --- & 87.4 \\
    SDAN~\cite{sdan} & 94.3 & {86.5} & {86.9} & \underline{95.1} & {91.1} & {90.0} & {82.1} & {77.9} & {96.4} & {77.2} & 86.6 & \underline{88.0} & {87.7}\\
    UFAL~\cite{ringwald2020unsupervised} & 94.9 & \underline{87.0} & 87.0 & {96.5} & \underline{91.8} & \textbf{95.1} & 76.8 & \textbf{78.9} & \underline{96.5} & \underline{80.7} & \underline{93.6}& 86.5 & \underline{88.8}\\\hline\Tstrut
    \textbf{CVP (ours)} & \underline{95.2} & \textbf{89.1} & \underline{91.2} & 94.6 & \textbf{92.4} & \underline{91.3} & \underline{82.2} & \underline{78.4} & \textbf{96.6} & \textbf{87.9} & 91.3 & \textbf{90.5} & \textbf{90.1}\\
\toprule
\end{tabular}
\end{adjustbox}
\caption{\label{table:visda_sota}Per class accuracy (in \%) for different methods on the \textbf{VisDA-2017 test set} as per challenge evaluation protocol. The domains are synthetic (source) and real (target).}
\end{table*}

\begin{table*}[!t]
\centering
\resizebox{0.99\textwidth}{!}{
\begin{tabular}{l@{\extracolsep{4pt}}rrrrrrr}
\toprule
    Method & P$\rightarrow$R & P$\rightarrow$S & R$\rightarrow$P & R$\rightarrow$S & S$\rightarrow$P & S$\rightarrow$R & Avg. \\\toprule
    Source only & \z{60.6}{1.2} & \z{28.7}{1.6} & \z{76.9}{0.4} & \z{26.9}{1.1} & \z{21.2}{2.4} & \z{8.8}{4.6} & 37.2 \\
    RSDA-DANN~\cite{gu2020spherical,ganin2014unsupervised} & \z{78.6}{0.1} & \z{48.5}{0.7} & \z{90.0}{0.0} & \z{43.9}{0.9} & \z{63.2}{1.1} & \z{37.0}{1.2} & 60.2\\
    RSDA-MSTN~\cite{gu2020spherical,mstn} & \z{73.8}{0.2} & \z{59.2}{0.2} & \z{87.5}{0.2} & \z{50.3}{0.5} & \z{69.5}{0.6} & \z{44.6}{1.1} & 64.2\\
    SymNets~\cite{zhang2019domain} & \z{81.4}{0.3} & \z{53.1}{0.6} & \textbf{\z{92.3}{0.2}} & \z{49.2}{1.0} & \z{69.6}{0.5} & \z{44.9}{1.5} & 65.1\\
    CAN~\cite{can} & \underline{\z{81.5}{0.3}} & \textbf{\z{72.3}{0.3}} & \underline{\z{92.2}{0.2}} & \underline{\z{69.3}{0.9}} & \underline{\z{74.9}{1.1}} & \underline{\z{60.7}{0.6}} & \underline{75.2}\\\hline\Tstrut
    \textbf{CVP (ours)} & \textbf{\z{81.7}{0.1}} & \underline{\z{71.6}{0.4}} & \z{91.5}{0.1} & \textbf{\z{71.8}{0.8}} & \textbf{\z{77.2}{1.7}} & \textbf{\z{62.3}{1.3}} & \textbf{76.0}\\
\toprule
\end{tabular}
}
\caption{Classification accuracy (in \%) for the \textbf{Adaptiope} dataset using a ResNet-50 backbone architecture. The domains are synthetic (S), real (R) and product (P).}\label{table:adaptiope_baselines}
\end{table*}

\textbf{Comparison to State-of-the-Art}. We now compare our proposed CVP algorithm to recently proposed unsupervised domain adaptation methods. On the VisDA-2017 dataset (see Table~\ref{table:visda_sota}), CVP is able to outperform recent methods and achieves the best accuracy for 5 out of 12 classes and improves the mean class average by 1.3\% \wrt to the previous SOTA result. To the best of our knowledge, this poses the best published result for the commonly used ResNet-101 backbone. Given the current VisDA-2017 challenge leaderboard~\cite{visdaleaderboard}, our CVP method would rank 2\textsuperscript{nd} place -- only behind a much larger $5\times$ResNet-152 ensemble that also averages results over multiple runs with test time augmentations. This, however, is not a fair comparison to our single ResNet-101 setup without such test time tricks.
\newline
On the Adaptiope dataset (see Table~\ref{table:adaptiope_baselines}), we compare to the recent CVPR proposals SymNets~\cite{zhang2019domain}, CAN~\cite{can} and RSDA~\cite{gu2020spherical} (using both DANN~\cite{ganin2014unsupervised} and MSTN~\cite{mstn}). Again, CVP is able to surpass these state-of-the-art algorithms and achieves the best accuracy on 4 out of 6 transfer tasks and also the overall average. 
Even for the hardest transfer tasks -- i.e.\ those involving synthetic images as source domain -- CVP manages to improve the results of CAN~\cite{can} by 1.6\% for S$\rightarrow$R and 2.3\% for S$\rightarrow$P. Again, CVP's overall accuracy average of 76\% poses the best published result and shows that our method can reliably conduct domain adaptation, even when confronted with Adaptiope's 123 difficult classes.
\newline
Furthermore, we report results on the Office-Caltech dataset in Table~\ref{table:officecaltech}. Again, our CVP method is able to achieve the best or second best results in 9 out of 12 transfer tasks and is also within 0.6\% of UFAL~\cite{ringwald2020unsupervised}. Please note that due to Office-Caltech's average target domain size of only 633 images, a 0.6\% difference arises from less than 2 misclassifications per transfer task on average. Therefore, our CVP method can be considered on par with the top performing methods UFAL~\cite{ringwald2020unsupervised}, PrDA~\cite{hua2020unsupervised} and RWA~\cite{rwa}.
\newline
Finally, we report results on the Modern Office-31 dataset in Table~\ref{table:modern_office}. Yet again, CVP is able to achieve the best or second best result in 4 out of the 6 available transfer tasks and is on par with the recently proposed RSDA~\cite{gu2020spherical} and CAN~\cite{can} methods. For the difficult A\textsubscript{ref}$\rightarrow$S task, CVP even manages to improve upon prior methods by 2.0\%. 
We thus conclude that CVP is able to successfully conduct unsupervised domain adaptation on a wide variety of transfer tasks and datasets and is able to achieve results at or close to state-of-the-art level.

\begin{table*}[t]
\centering
\resizebox{1.0\textwidth}{!}{
\begin{tabular}{l@{\extracolsep{4pt}}rrrrrrrrrrrrr}
\toprule
    \multirow{2}{*}{Method} & \multicolumn{3}{c}{A} & \multicolumn{3}{c}{C} & \multicolumn{3}{c}{D} & \multicolumn{3}{c}{W} & \multirow{2}{*}{Avg.}\\\cline{2-4}\cline{5-7}\cline{8-10}\cline{11-13}
     & C & D & W & A & D & W & A & C & W & A & C & D & \\\toprule
    CORAL \cite{sun2016return,rwa} & 89.2 & 92.2 &  91.9 &  94.1 &  92.0 &  92.1  & 94.3 &  87.7 &  98.0 &  92.8 &  86.7 & \textbf{100.0} & 92.6\\
    GTDA+LR \cite{vascon2019unsupervised}  & 91.5 & 98.7 & 94.2 & 95.4 & \underline{98.7} & 89.8 & 95.2 & 89.0 & 99.3 & 95.2 & 90.4 & \textbf{100.0} & 94.8\\
    RWA \cite{rwa} & 93.8 & {98.9} & {97.8} & 95.3 & \textbf{99.4} & 95.9 & 95.8 & 93.1 & 98.4 & 95.3 & 92.4 & {99.2} & 96.3\\
    PrDA \cite{hua2020unsupervised} & 92.1 & \underline{99.0} & \underline{99.3} & \textbf{97.2} & \textbf{99.4} & \underline{98.3} & 94.7 & 91.0 & \underline{99.7} & 95.6 & 93.4 & \textbf{100.0} & 96.6 \\
    UFAL~\cite{ringwald2020unsupervised} & \textbf{95.1} & \textbf{99.4} & \textbf{99.7} & {96.0} & 96.8 & \textbf{99.7} & \underline{95.8} & \textbf{95.0} & \underline{99.7} & \textbf{96.3} & \underline{95.0} & \underline{99.4} & \textbf{97.3} \\\hline\Tstrut
    \textbf{CVP (ours)} & \underline{94.3}  & 98.1 & 99.0 & \underline{96.6} & 91.7 & \underline{98.3} & \textbf{95.9} & \underline{94.7} & \textbf{100.0} & \underline{96.1} & \textbf{95.3} & \textbf{100.0} & \underline{96.7} \\
\toprule
\end{tabular}}
\caption{\label{table:officecaltech}Classification accuracy (in \%) for different methods using ResNet-50 on the \textbf{Office-Caltech} dataset with domains Amazon, Caltech, DSLR and Webcam.}
\end{table*}

\renewcommand{\z}[2]{#1}
\begin{table}[!t]
\centering
\resizebox{0.8\textwidth}{!}{
\begin{tabular}{l@{\extracolsep{4pt}}rrrrrrr}
\toprule
    Method & A\textsubscript{ref}$\rightarrow$S & A\textsubscript{ref}$\rightarrow$W & W$\rightarrow$A\textsubscript{ref} & W$\rightarrow$S & S$\rightarrow$A\textsubscript{ref} & S$\rightarrow$W & Avg. \\\toprule
    Source only~\cite{ringwald2021adaptiope} & \z{51.3}{0.4} & \z{76.8}{1.0} & \z{74.1}{0.5} & \z{51.3}{1.0} & \z{14.9}{3.5} & \z{8.0}{1.5} & \z{46.1}{1.3}\\
    SymNet~\cite{zhang2019domain} & \z{65.9}{1.0} & \z{91.0}{0.2} & \z{89.2}{0.4} & \z{56.5}{1.4} & \z{86.8}{2.1} & \z{82.2}{1.2} & \z{78.6}{0.6}\\
    RSDA-DANN~\cite{gu2020spherical,ganin2014unsupervised} & \z{76.1}{0.5} & \z{91.8}{0.5} & \z{90.5}{0.9} & \z{70.4}{1.0} & \z{80.8}{0.3} & \z{83.1}{2.8} & \z{82.1}{0.7}\\
    RSDA-MSTN~\cite{gu2020spherical,mstn} & \underline{\z{82.0}{0.6}} & \underline{\z{92.2}{0.3}} & \textbf{\z{93.0}{0.6}} & \z{76.3}{0.7} & \z{90.0}{1.5} & \underline{\z{86.2}{2.9}} & \underline{\z{86.6}{0.3}}\\
    CAN~\cite{can} & \z{79.1}{3.0} & \textbf{\z{92.8}{0.5}} & \z{90.9}{1.0} & \textbf{\z{77.9}{1.6}} & \textbf{\z{91.2}{0.5}} & \textbf{\z{89.7}{0.5}} & \textbf{\z{86.9}{0.7}}\\\hline\Tstrut
    \textbf{CVP (ours)} & \textbf{84.0} & 91.4 & \underline{91.1} & \underline{77.5} & \underline{90.4} & 85.2 & \underline{86.6} \\
\toprule
\end{tabular}}
\caption{Classification accuracy (in \%) on the \textbf{Modern Office-31} dataset using ResNet-50.}\label{table:modern_office}
\end{table}

\section{Conclusion}

In this paper, we propose CVP -- the Certainty Volume Prediction algorithm -- for estimation of uncertainty in feature space. We show how CVP's sampling stage interacts with the decision boundary and how this can improve a model's generalization properties and therefore domain adaptation capabilities. Additionally, we show that CVP's $\sigma$ estimation in feature space relates to other common uncertainty quantifications in classifier space. Due to the intuitive interpretability of $\sigma$, we also see great potential for other downstream tasks, \eg employing $\sigma$ for out-of-distribution rejection in open-set domain adaptation.
Overall, we demonstrate that making a model uncertainty-aware through CVP can improve its predictions and achieve state-of-the-art results on challenging benchmark UDA datasets.

\setlength\cellspacetoplimit{10pt}
\setlength\cellspacebottomlimit{10pt}
\newcommand\cincludegraphics[2][]{\raisebox{-0.3\height}{\includegraphics[#1]{#2}}}

\setcounter{section}{0}
\setcounter{table}{0}
\setcounter{figure}{0}

\renewcommand{\thetable}{\Roman{table}}
\renewcommand{\thefigure}{\Roman{figure}}

\appendix
\section*{Appendices}
\addcontentsline{toc}{section}{Appendices}
\renewcommand{\thesubsection}{\Alph{subsection}}
\section{Implementation Details} \label{sec:appendix_implementation}

In this section, we provide further detail about the employed setup and evaluation.
For our experiments on VisDA-2017~\cite{visda2017}, we employ the common ResNet-101~\cite{he2016deep} feature extractor ($f_\text{CNN}$) similar to \cite{can,french2017self,ringwald2020unsupervised}, while for Office-Caltech~\cite{officecaltech}, Modern Office-31~\cite{ringwald2021adaptiope} and Adaptiope~\cite{ringwald2021adaptiope}, we use ResNet-50 as proposed in~\cite{ringwald2021adaptiope}. In all cases, $f_\text{CNN}$ is pretrained on ImageNet~\cite{imagenet}. We implement the $f_\Sigma$ subnetwork as a two layer linear network. The input size is 2048 (same size as $\mu$), hidden size 2048 and output size 1. Intermediate outputs are activated by ReLU. The final output is activated by the softplus function in order to enforce $\sigma \in \mathbb{R}^+_0$ and the positive semi-definiteness of $\Sigma$.
\newline
We jointly optimize all parameters using SGD with Nesterov momentum~\cite{nesterov} of 0.95 and a learning rate of $5\times 10^{-4}$, which is decreased according to the schedule used in~\cite{can}. 
Networks are trained for 250 cycles with $T_\text{cycle}=50$ forward passes each. 
For the evaluation of the decision boundary, we set $K=1000$ in order to sample 1000 subfeatures (including start and end feature).
Our code is implemented in PyTorch~\cite{paszke2017automatic} and will be made available. 
\newline
\textbf{Reparameterization Trick}. Neural network training is based on the idea of backpropagation and therefore requires the whole network to be differentiable. This, however, is not the case when including a generally non-differentiable step such as sampling from a distribution. In this work, we want the parameters of $f_\Sigma$ to also be updated based on our proposed $\mathcal{L}_\text{ce}^\varphi$ loss function. We thus sidestep the non-differentiability of sampling by applying the reparameterization trick proposed in~\cite{kingma2013auto}. Here, instead of sampling a feature $z$ directly from $z \sim \mathcal{N}(\mu, \Sigma)$, we first draw an auxiliary noise variable $\varepsilon \sim \mathcal{N}(\mathbb{0}, \mathbb{1})$ and then compute $z=\mu + \varepsilon \Sigma^\frac{1}{2}$. This can also be rewritten as $\forall i: z_i=\mu_i+\varepsilon_i\; \mathrm{diag}(\Sigma^\frac{1}{2})_{i}$ with $\varepsilon_i \sim \mathcal{N}(0, 1)$ (note the scalar notation). Applying this reparameterization allows the gradient to also flow through our $f_\Sigma$ subnetwork.

\section{Theorem for $\kappa$ Initialization} \label{sec:kappa}

We now derive the constant $\kappa$ (used in the $\psi^{(i)}$ equation in the main paper) as the expected cross-entropy value given a randomly initialized classifier.
Consider the cross-entropy term $H(q, p) = -\sum q\; \mathrm{log}(p)$ between two categorical distributions $q$ and $p$ with $\lvert \mathcal{C} \rvert$ classes. Given an uniform class distribution, the one-hot encoded target $q_i$ will equal 1 at position $i$ with probability $\frac{1}{\lvert \mathcal{C} \rvert}$. We can then rewrite $H(q, p) = -\sum_{i=1}^{\lvert \mathcal{C} \rvert} \frac{1}{\lvert \mathcal{C} \rvert}\; \mathrm{log}(p_i)$. This can be further simplified as Equation~\ref{eq:simplified}.

\begin{align}
    H(q, p) = -\frac{1}{\lvert \mathcal{C} \rvert} \sum_{i=1}^{\lvert \mathcal{C} \rvert} \mathrm{log}(p_i) = -\frac{1}{C} (\mathrm{log}(p_1) + ... + \mathrm{log}(p_{\lvert \mathcal{C} \rvert})) \label{eq:simplified}
    = -\frac{1}{\lvert \mathcal{C} \rvert} \mathrm{log}(p_1\cdot p_2 \ldots \cdot p_{\lvert \mathcal{C} \rvert})
\end{align}

We can assume that the worst state of a model during training is to be equal to a random oracle, i.e. we assume that $\forall i: \mathbb{E}[p_i] \approx \frac{1}{\lvert \mathcal{C} \rvert}$ holds true. 
Given this assumption, we can rewrite Equation~\ref{eq:simplified} as $H(q, p) =  -\frac{1}{\lvert \mathcal{C} \rvert} \mathrm{log}(\frac{1}{\lvert \mathcal{C} \rvert}^{\lvert \mathcal{C} \rvert}) = -\frac{\lvert \mathcal{C} \rvert}{\lvert \mathcal{C} \rvert} \mathrm{log}(\frac{1}{\lvert \mathcal{C} \rvert}) = -\mathrm{log}(\frac{1}{\lvert \mathcal{C} \rvert})$, 
which further simplifies to $\mathrm{log}(\lvert \mathcal{C} \rvert)$. We thus define constant $\kappa = \mathrm{log}(\lvert \mathcal{C} \rvert)$, which can be seen as an expected upper bound for $\mathcal{L}_{ce}^{\varphi}$.

\section{Additional Analysis} \label{sec:appendix_analysis}

\begin{figure}[!t]
\renewcommand{\arraystretch}{2.0}
\begin{tabular}{ccc}
\bmvaHangBox{\includegraphics[width=0.3\textwidth]{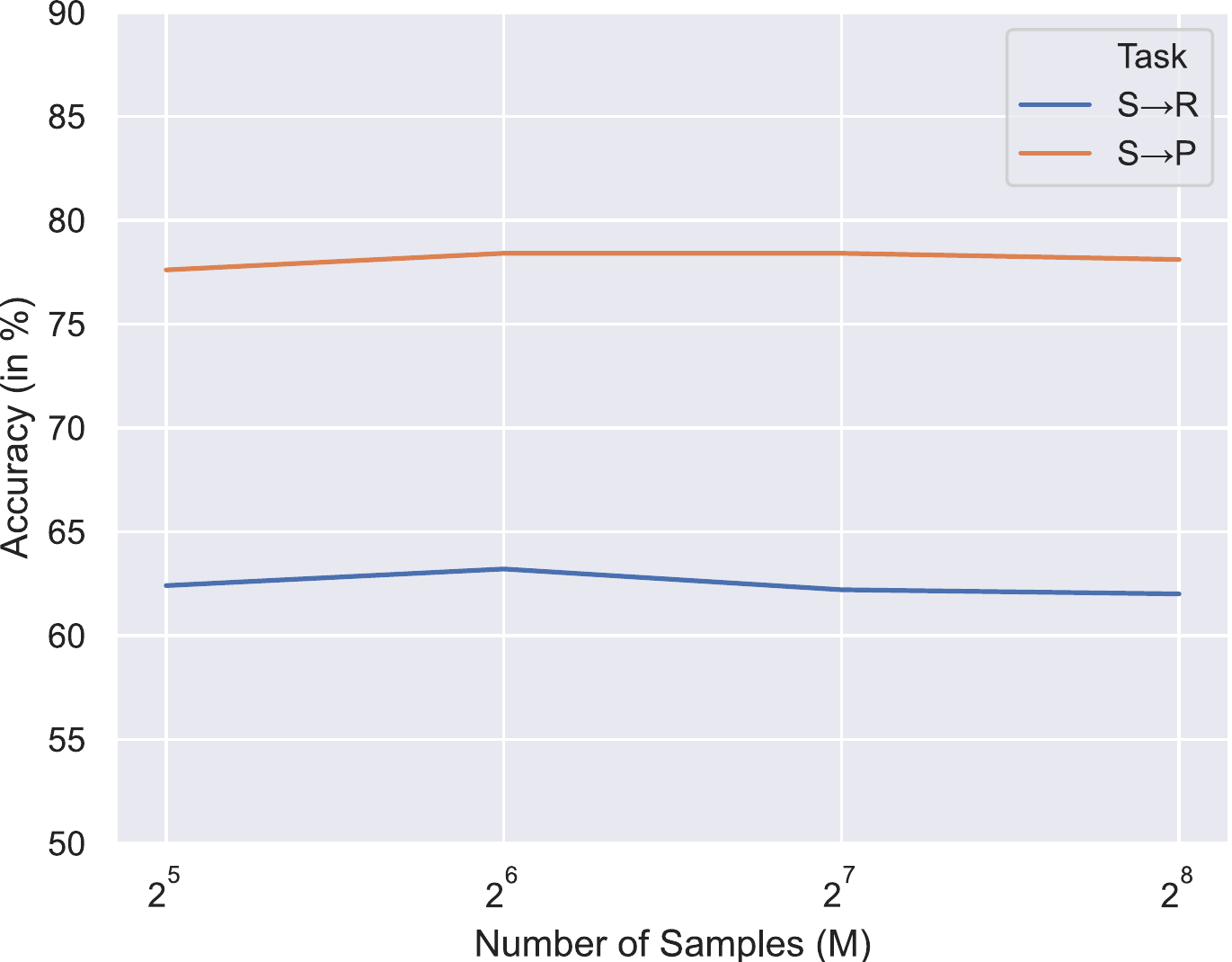}} & \bmvaHangBox{\includegraphics[width=0.3\textwidth]{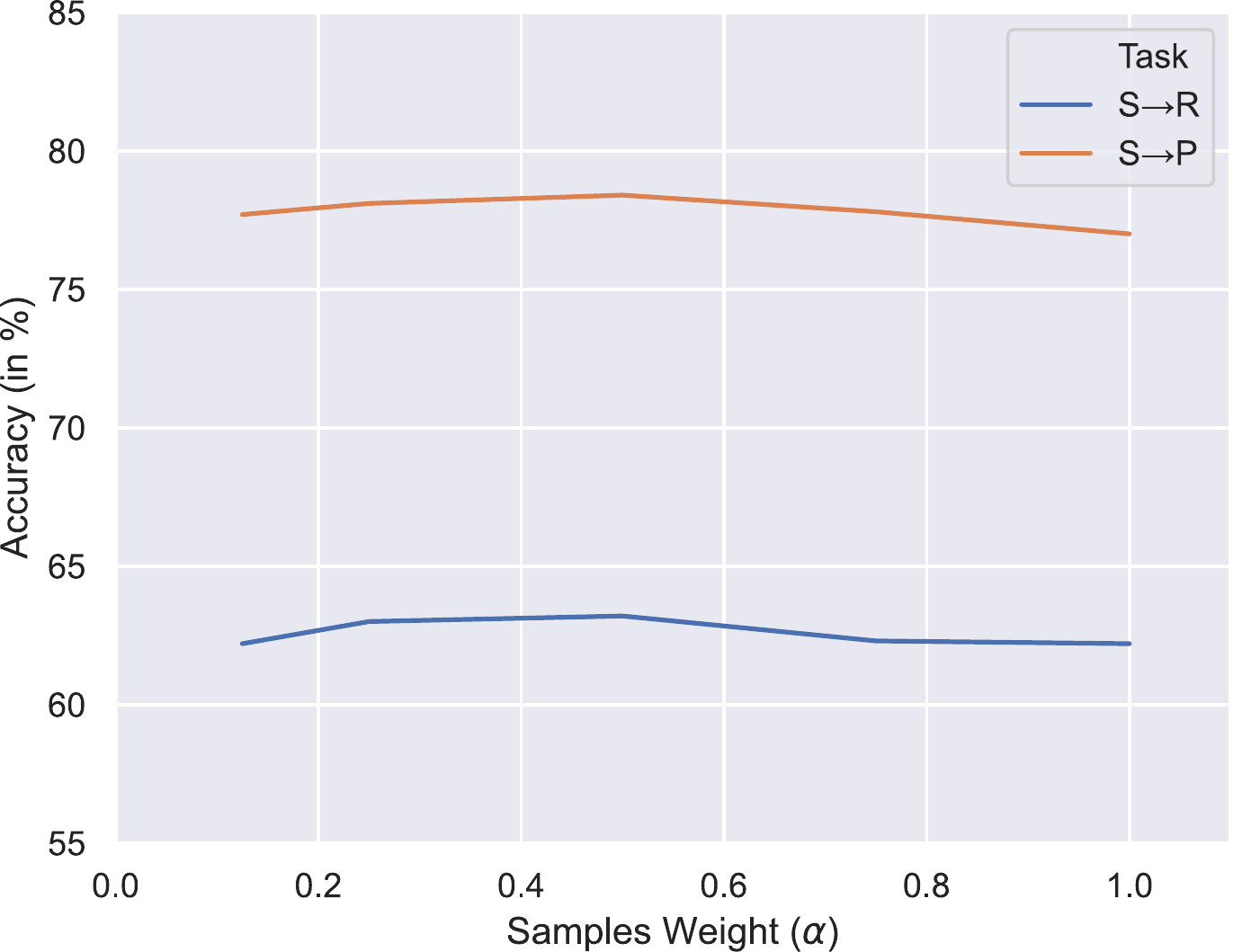}} &
\bmvaHangBox{\includegraphics[width=0.3\textwidth]{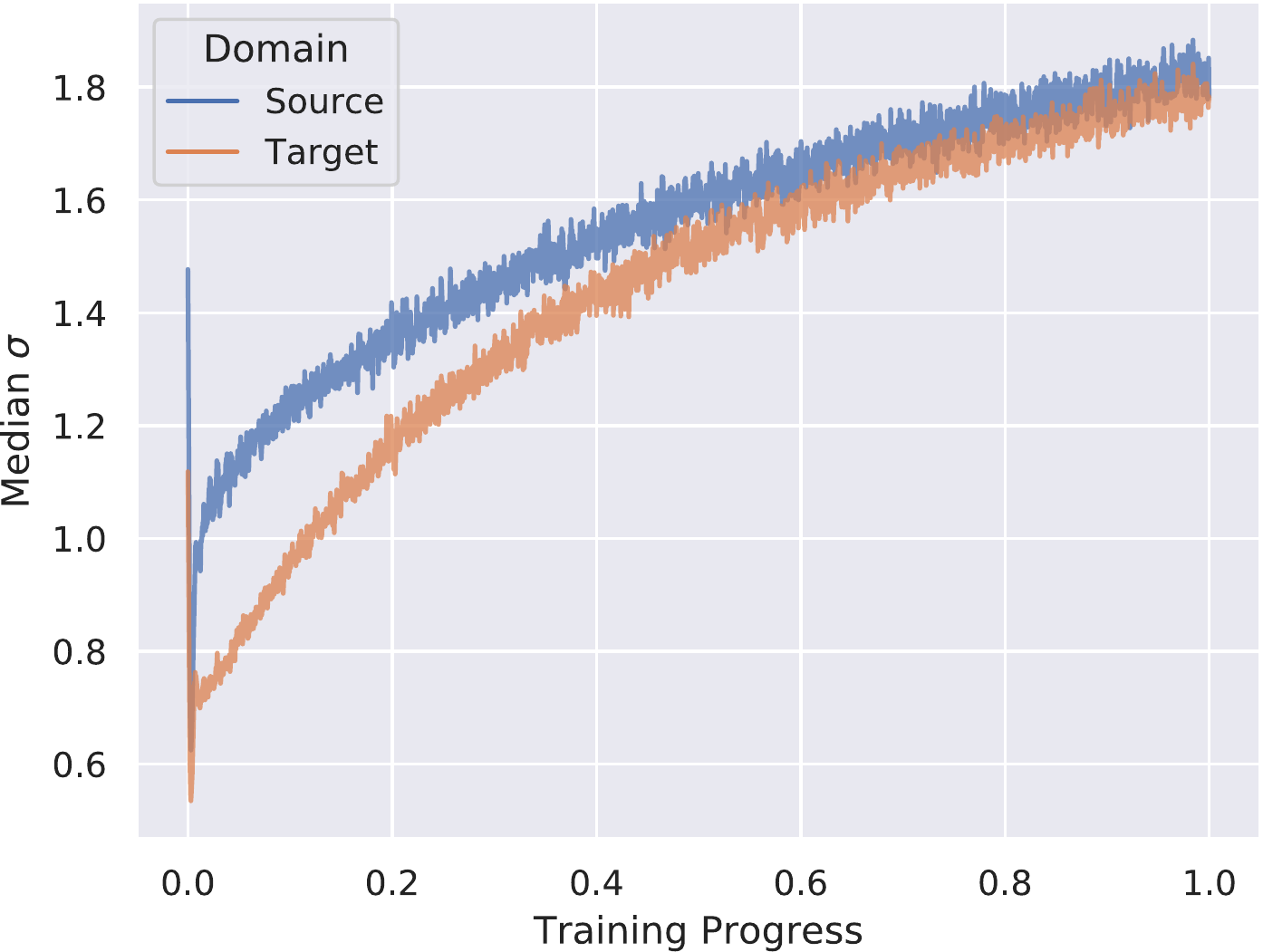}}\\
	(a) Number of Samples $M$ & (b) Samples Weight $\alpha$ & (c) P$\rightarrow$S \\
\end{tabular}
\caption{(a, b) Results for different ablation studies. (c) Median $\sigma$ values for the source and target domain as the training progresses. Best viewed in the digital version.} 
\label{fig:ablation}
\end{figure}

\begin{figure*}[!t]
\begin{tabular}{ Sc  Sc  Sc  Sc  Sc  Sc }
\centering
\cincludegraphics[width=0.2\textwidth]{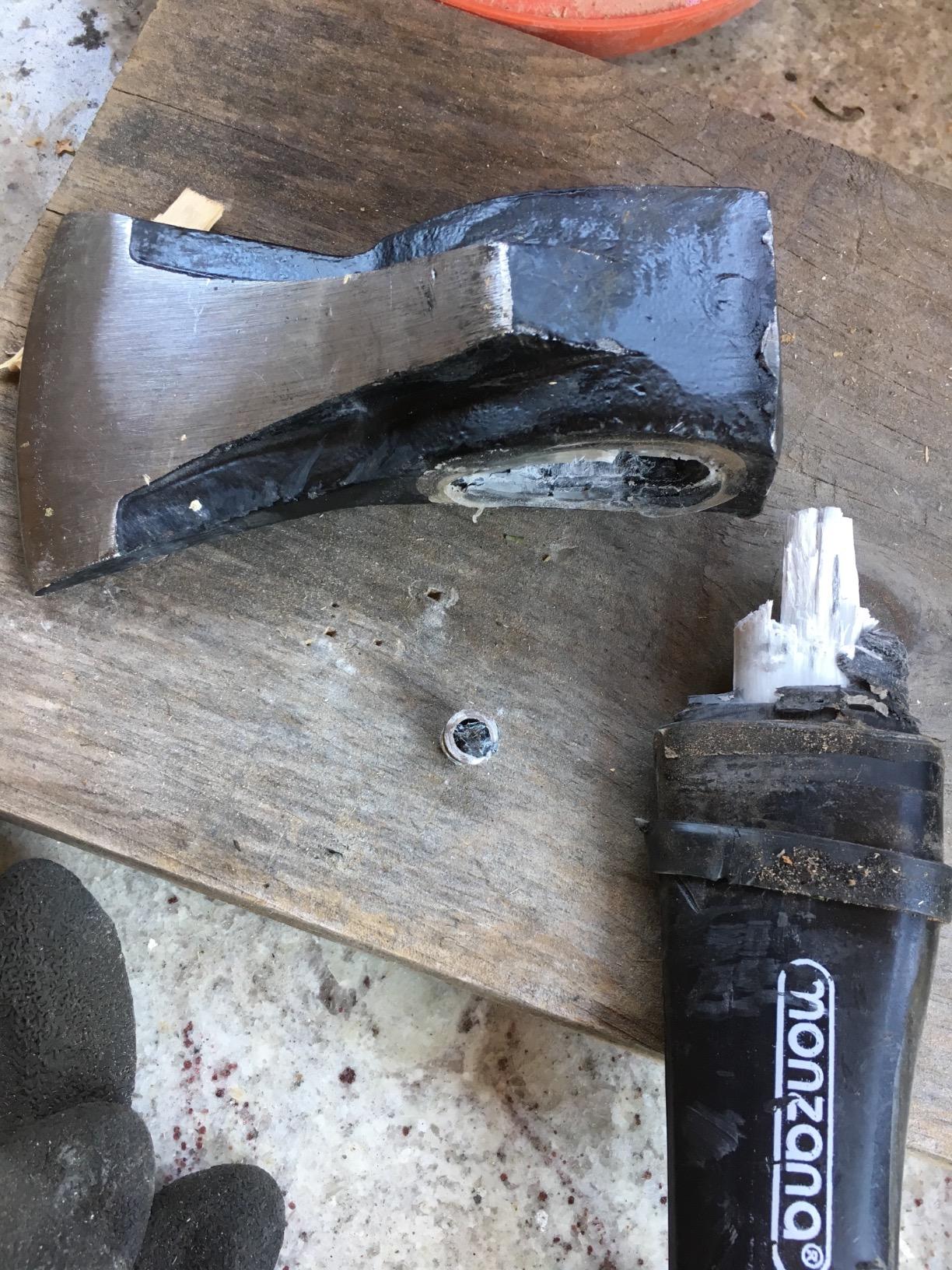} &
\raisebox{2.1cm}{\multirow{3}{*}{\cincludegraphics[width=0.2\textwidth, height=8.6cm]{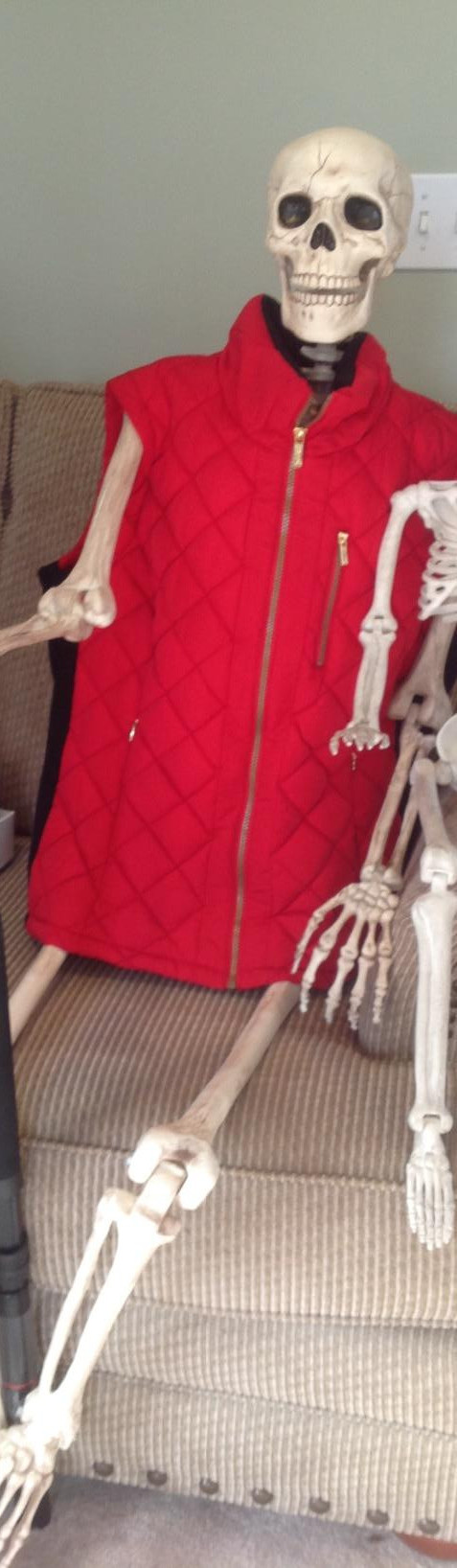}}} && & \cincludegraphics[width=0.2\textwidth]{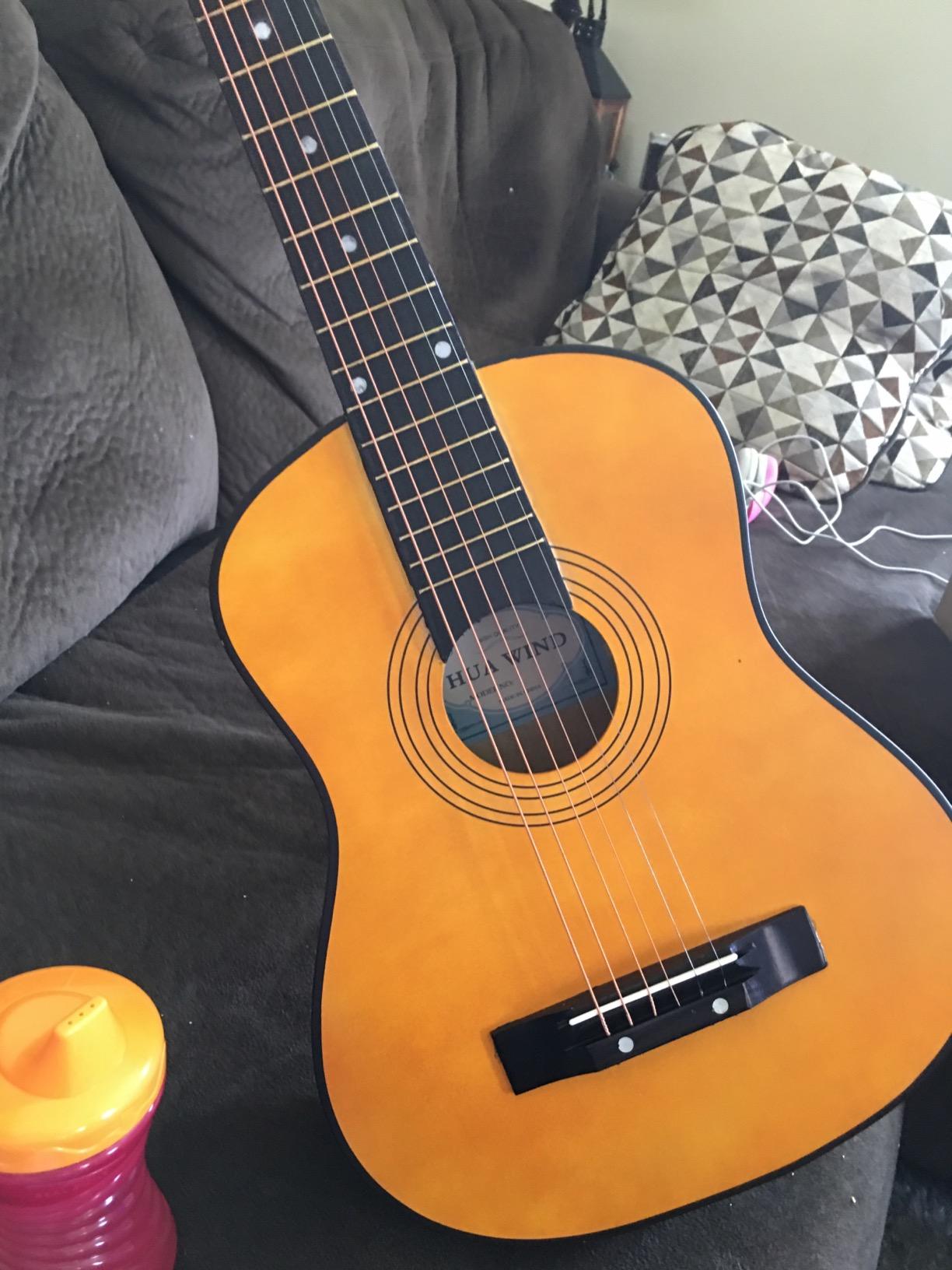} & \cincludegraphics[width=0.2\textwidth]{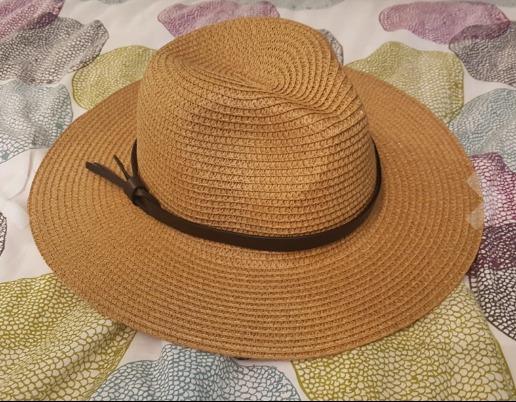}\\
(a) axe & && & (b) acoustic guitar & (c) hat \\
\cincludegraphics[width=0.2\textwidth]{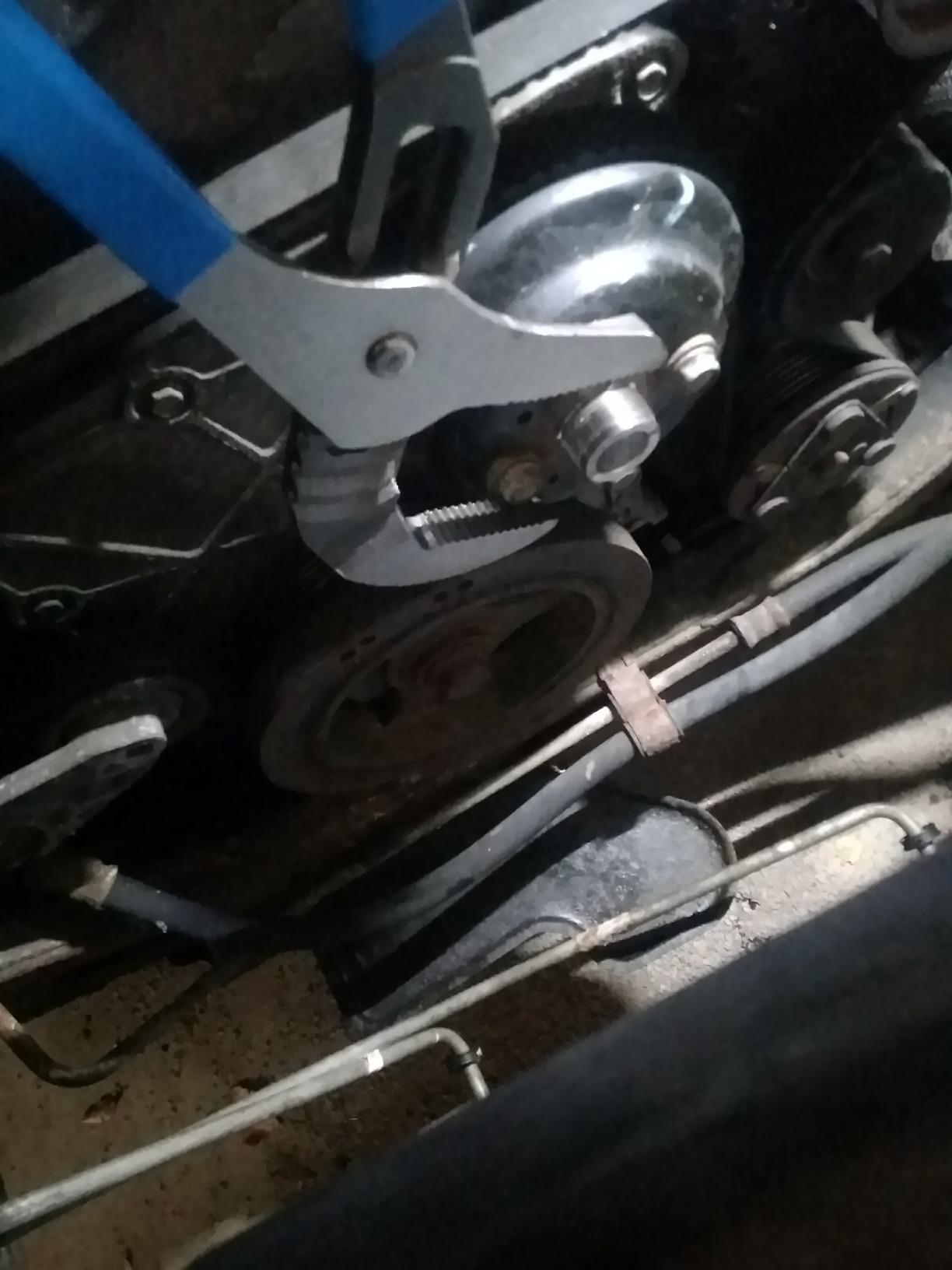} & && & \cincludegraphics[width=0.2\textwidth]{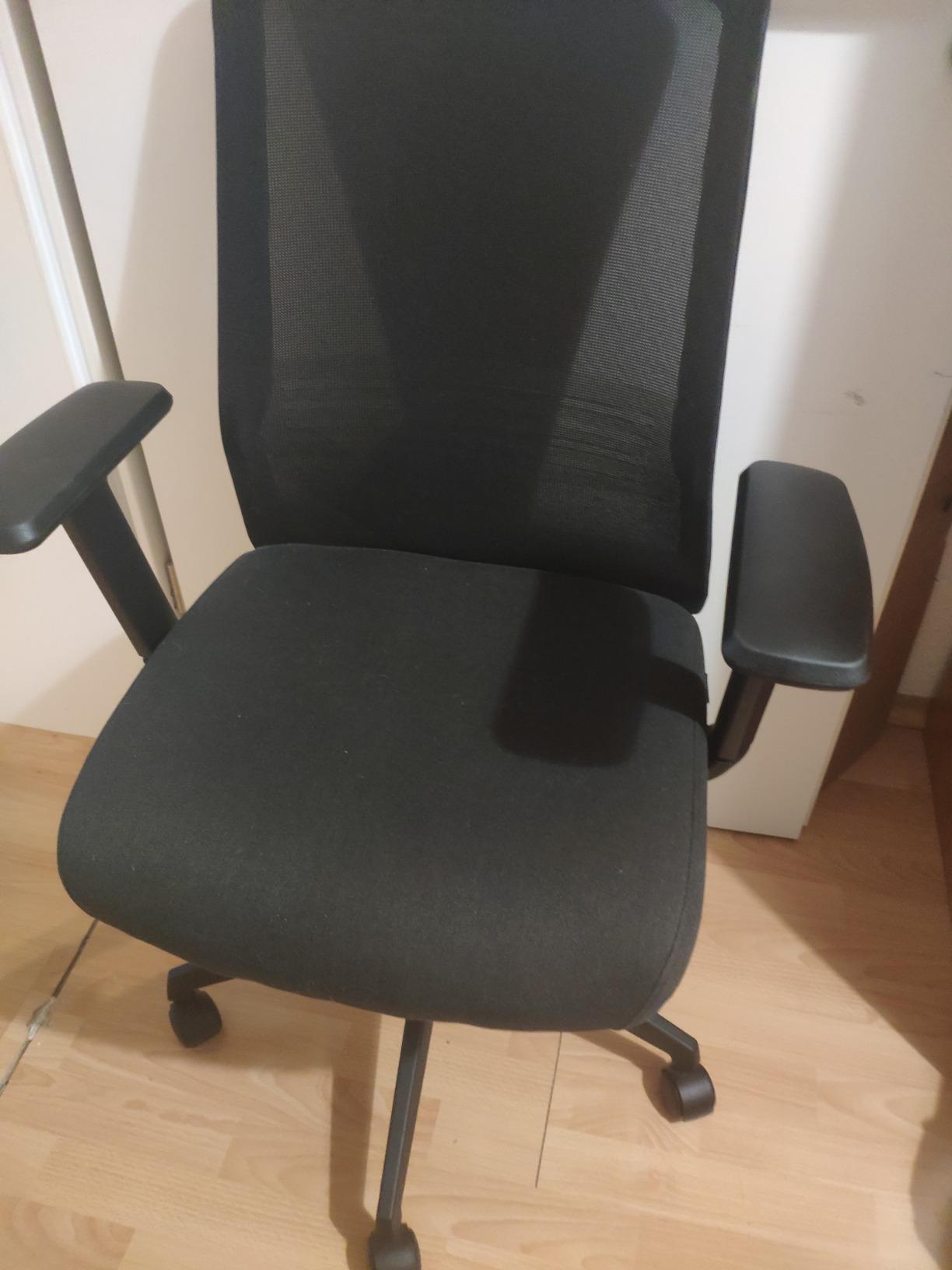} & \cincludegraphics[width=0.2\textwidth]{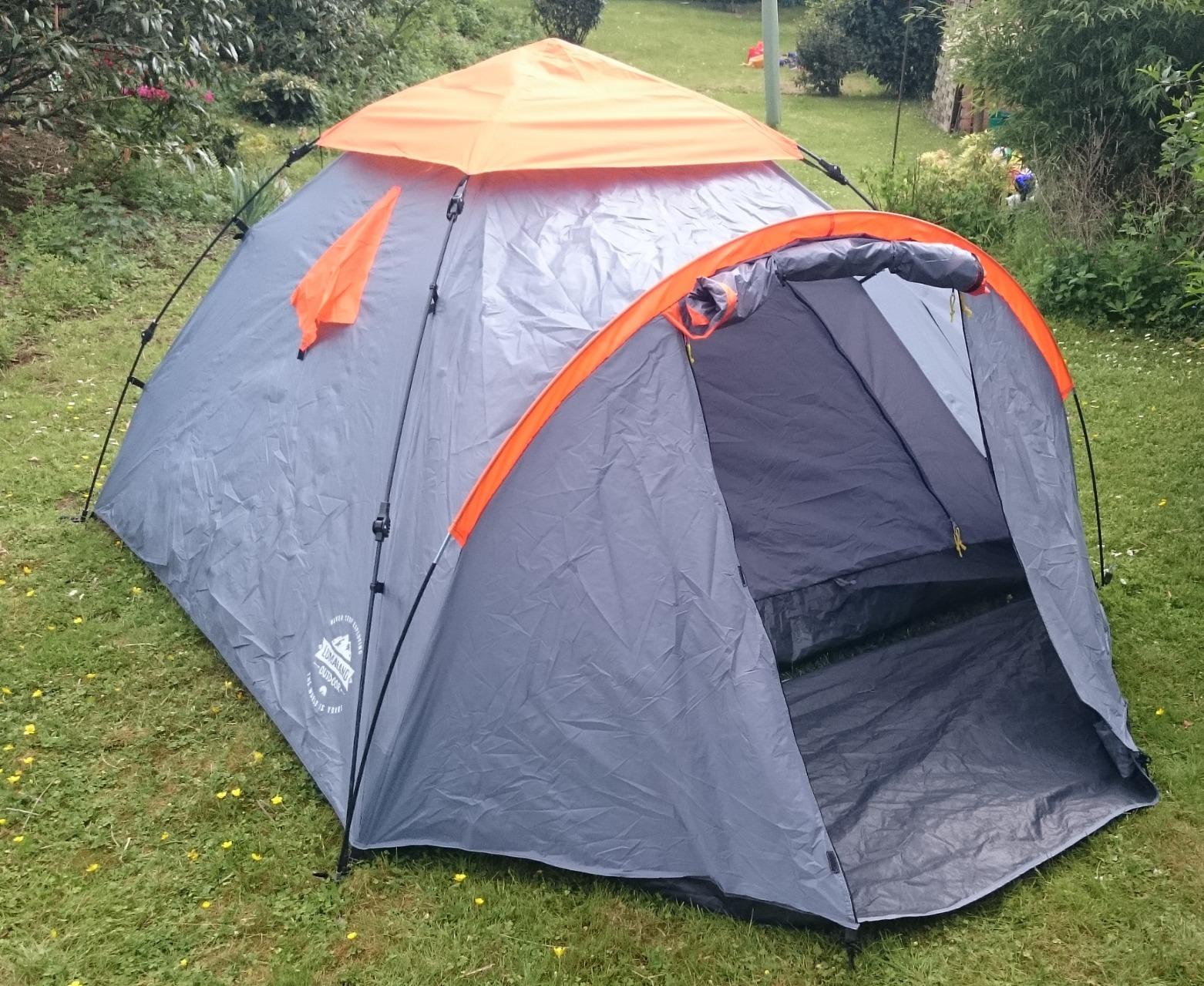}\\
(d) pipe wrench & (e) skeleton && & (f) office chair & (g) tent \\
\multicolumn{3}{c}{\textbf{Low $\sigma$ magnitude}} & \multicolumn{3}{c}{\textbf{High $\sigma$ magnitude}} \\
\end{tabular}
\caption{Left two columns: Images associated with a low magnitude of $\sigma$. Right two columns: Images associated with high magnitudes of $\sigma$. Image captions denote the ground truth class.} 
\label{fig:interpret}
\end{figure*}

\textbf{Ablation Study}. We also evaluate our proposed CVP method by conducting an ablation study \wrt its two hyperparameters $\alpha$ and $M$. For this evaluation, we use Adaptiope's S$\rightarrow$P and S$\rightarrow$R transfer tasks. Hyperparameter $M$ controls the amount of samples that are drawn from the distribution predicted by CVP. We analyze $M$ in the range $[2^5, 2^8]$ and plot the results in Figure~\ref{fig:ablation}a. We notice that CVP is fairly robust \wrt to this parameter as long as enough samples are drawn. Overall $M=2^6$ resulted in the best performance. 
Similarly, we analyze the samples weight $\alpha$ in Figure~\ref{fig:ablation}b. We observe that choosing $\alpha$ either to low or too high degrades the accuracy consistently for both transfer tasks. Intuitively, a too high $\alpha$ would attribute the generated samples the same weight as the predicted mean feature $\mu$, which can lead to optimization problems, as the samples are randomly drawn and not an actual output of the feature extractor. Related to this, setting $\alpha$ too low removes the additional supervision to enforce the cluster assumption. We thus set $\alpha=0.5$ for our experiments, due to its overall best performance. 
\newline
\textbf{Sigma Analysis}. Furthermore, we provide an additional analysis for the P$\rightarrow$S task of Adaptiope similar to the analysis in the main paper in Figure~\ref{fig:ablation}c. We note the same key observations as for the S$\rightarrow$P and S$\rightarrow$R tasks in the main paper: Directly after the switch from the pretraining to the adaptation phase, $\sigma$ drops harshly due to being confronted with new, unseen target domain data. For the source domain, $\sigma$ recovers quickly while for the target domain, $\sigma$ grows slower due to the ongoing domain adaptation. Unlike the S$\rightarrow\ast$ tasks, the source and target domain $\sigma$ end up very close to each other for the S$\rightarrow$P task. This, however, is expected due to the smaller domain gap and the missing background clutter: both the synthetic and the product domain have cropped items with either a fully white (product domain) or fully grey (synthetic domain) background. 
\newline
\textbf{Interpretability of $\sigma$}. We further analyze the interpretability of $\sigma$. For this, we run our full CVP method on Adaptiope's S$\rightarrow$R task and extract $\sigma$ values for the whole target domain. In Figure~\ref{fig:interpret}, we show selected results from the top and bottom 20 $\sigma$ values.
Images associated with high magnitudes of $\sigma$ are usually taken under good conditions \wrt lighting and camera angle and commonly represent the best images of their respective object classes. 
For low magnitudes of $\sigma$, we often find objects to be in unusual positions (see Figure~\ref{fig:interpret}d), disassembled or broken (see Figure~\ref{fig:interpret}a) or in exceptionally uncommon conditions or situations (see Figure~\ref{fig:interpret}e). This further reinforces the usability of $\sigma$ as an interpretable uncertainty measure.

\clearpage

\section{Visualization}
\begin{figure}[!h]
\centering
\renewcommand{\arraystretch}{2.0}
\begin{tabular}{cc}
\bmvaHangBox{\includegraphics[width=0.45\textwidth]{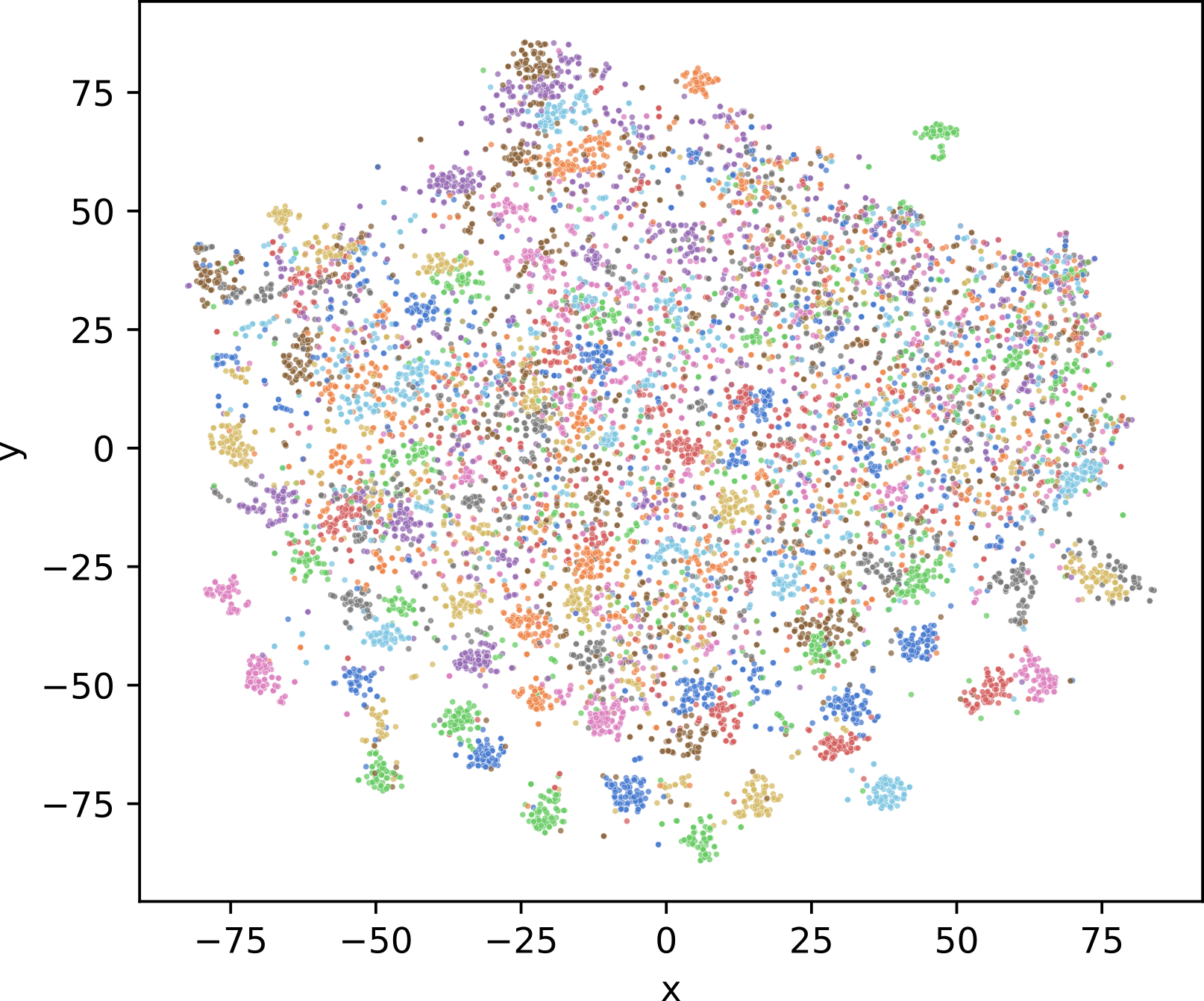}} & \bmvaHangBox{\includegraphics[width=0.45\textwidth]{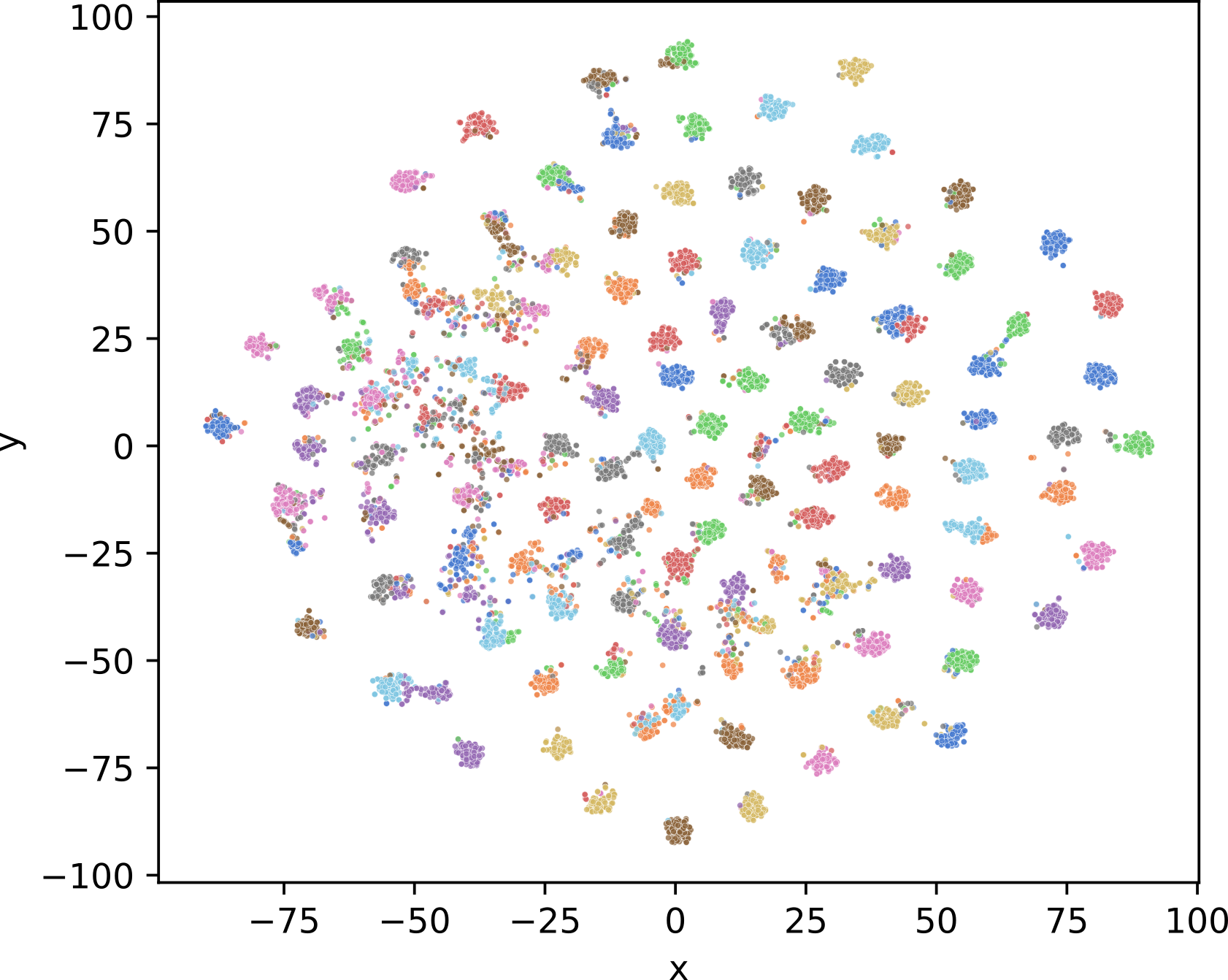}} \\
\bmvaHangBox{\includegraphics[width=0.45\textwidth]{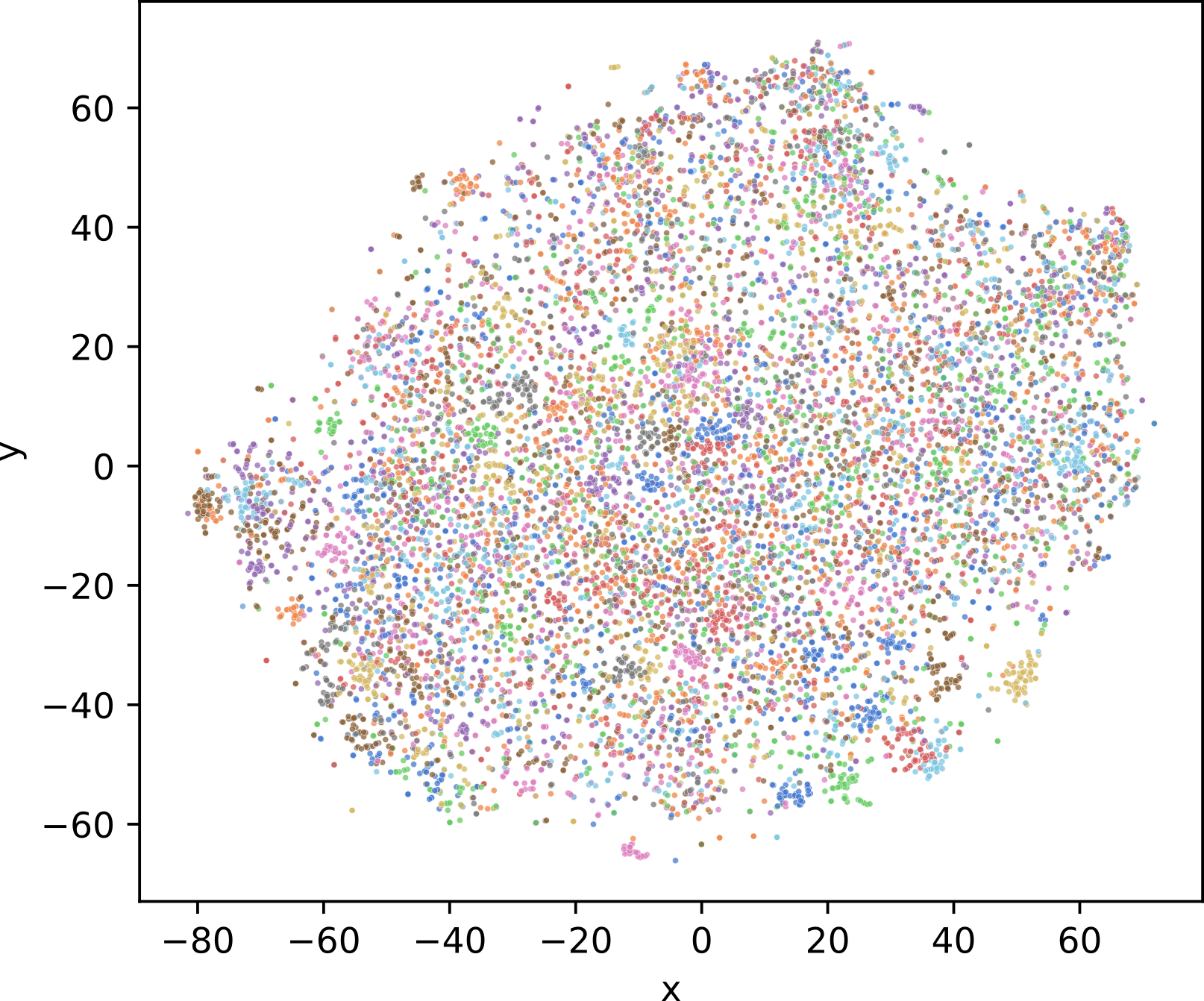}} & \bmvaHangBox{\includegraphics[width=0.45\textwidth]{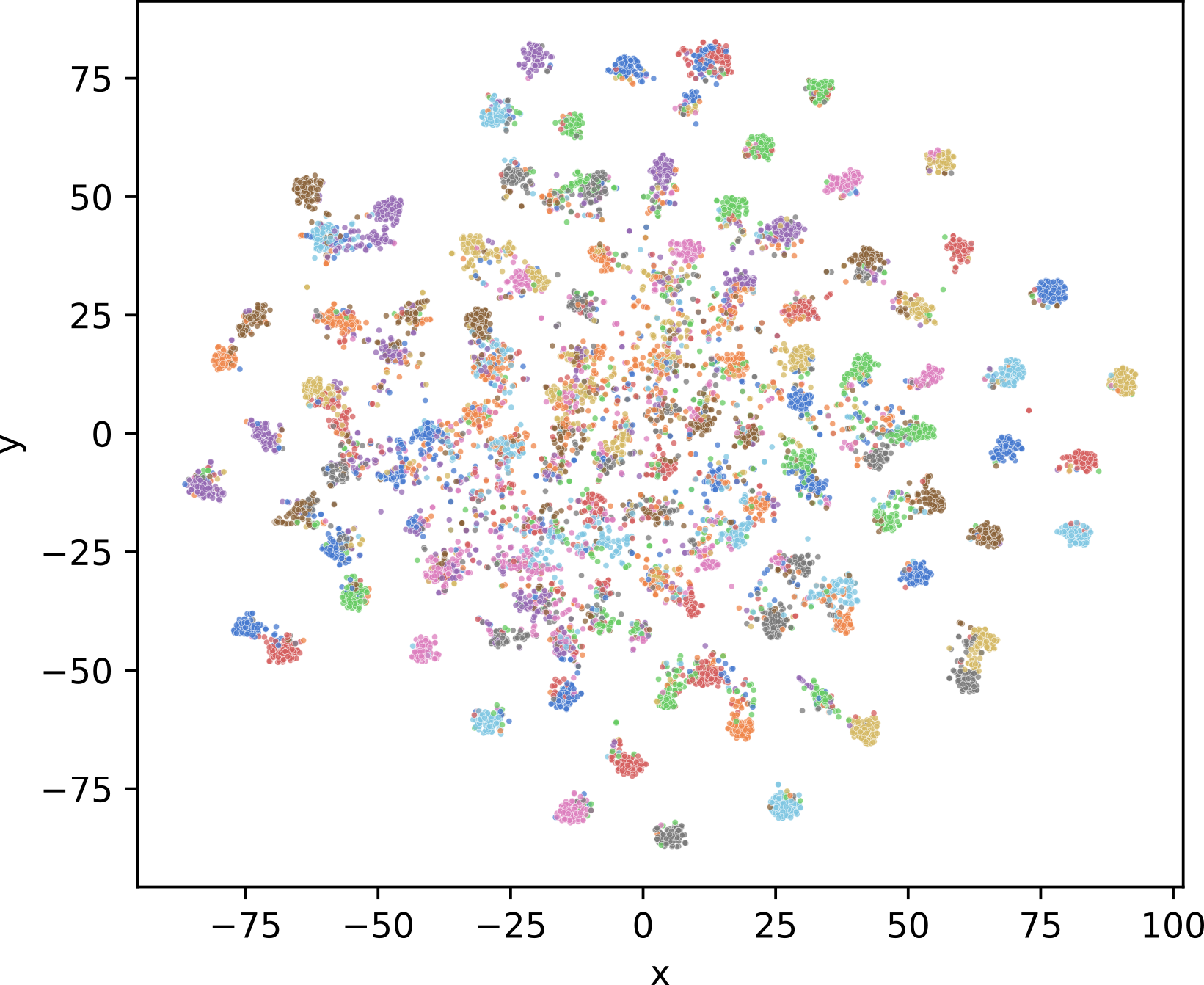}} \\
 (a) After Pretraining & (b) After Adaptation\\
\end{tabular}
\caption{Visualization of target domain features after the source-only pretraining and after the domain adaptation with CVP. Features were extracted for Adaptiope's S$\rightarrow$P (top row) and S$\rightarrow$R (bottom row) transfer tasks. Every color represents one of Adaptiope's 123 classes. Please note that due to the high number of classes, some colors are hard to distinguish.} 
\label{fig:visualization}
\end{figure}

In Figure~\ref{fig:visualization}, we provide a feature visualization for the features learned by our CVP method. For this, we extract the 2048-dimensional feature $\mu$ for every target domain image in Adaptiope's S$\rightarrow$P and S$\rightarrow$R transfer tasks respectively. The feature extraction step is conducted right after the source-only pretraining and after the finalized domain adaptation phase. We then employ t-SNE~\cite{maaten2008visualizing} in order to project the features into a two-dimensional space. As evident from Figure~\ref{fig:visualization}a, the features show no discernible structure with regard to clusters or proximity of same-class features after the source-only pretraining step. However, after the adaptation phase with CVP (Figure~\ref{fig:visualization}b), clear clusters of same-class features can be observed. We thus conclude that -- even for the very hard synthetic to real and synthetic to product transfer tasks of Adaptiope -- our CVP method was able to learn discriminative features even though no target domain labels were available.

\bibliography{main}
\end{document}